\documentclass[letterpaper,twocolumn,10pt]{article}
\usepackage{usenix2019_v3}
\usepackage{xr}
\usepackage{censor}

\usepackage[backend=biber,style=numeric-comp,sorting=none]{biblatex}
\usepackage{soul}
\usepackage{tikz}
\usepackage{floatrow}
\usepackage{multirow}
\usepackage{tikz}
\usepackage[skins]{tcolorbox}
\usepackage{amsmath}
\usepackage{cleveref}
\usepackage{cellspace}
\usepackage{url}
\usepackage{booktabs}
\usepackage{enumitem}
\usepackage{amssymb}
\usepackage{rotating}
\usepackage[T1]{fontenc}
\usepackage{xcolor} 
\usepackage{colortbl}
\usepackage{filecontents}

\newcommand*\fullcirc[1][1ex]{\tikz\fill (0,0) circle (#1);} 

\newcolumntype{L}[1]{>{\raggedright\let\newline\\\arraybackslash\hspace{0pt}}m{#1}}
\newcolumntype{C}[1]{>{\centering\let\newline\\\arraybackslash\hspace{0pt}}m{#1}}
\newcolumntype{R}[1]{>{\raggedleft\let\newline\\\arraybackslash\hspace{0pt}}m{#1}}

\usepackage{soul}
\DeclareRobustCommand{\hlcyan}[1]{{\sethlcolor{cyan!50}\hl{#1}}}
\DeclareRobustCommand{\hlred}[1]{{\sethlcolor{pink!80}\hl{#1}}}
\DeclareRobustCommand{\hlgreen}[1]{{\sethlcolor{green!50}\hl{#1}}}
\DeclareRobustCommand{\hlorange}[1]{{\sethlcolor{orange!50}\hl{#1}}}
\usepackage{tikz}
\newcommand*\circled[1]{\tikz[baseline=(char.base)]{
            \node[shape=circle,draw,inner sep=0.5pt] (char) {#1};}}

\begin{document}
%-------------------------------------------------------------------------------
\date{}

\title{\Large \bf SoK: Machine Learning for Misinformation Detection}

\author{
{\rm Madelyne Xiao}\\
Princeton University
\and
{\rm Jonathan Mayer}\\
Princeton University
} 

\maketitle

%-------------------------------------------------------------------------------

\begin{abstract}
%-------------------------------------------------------------------------------
We examine the disconnect between scholarship and practice in applying machine learning to trust and safety problems, using misinformation detection as a case study. We survey literature on automated detection of misinformation across a corpus of 248 well-cited papers in the field. We then examine subsets of papers for data and code availability, design missteps, reproducibility, and generalizability. Our paper corpus includes published work in security, natural language processing, and computational social science. Across these disparate disciplines, we identify common errors in dataset and method design. In general, detection tasks are often meaningfully distinct from the challenges that online services actually face. Datasets and model evaluation are often non-representative of real-world contexts, and evaluation frequently is not independent of model training. We demonstrate the limitations of current detection methods in a series of three representative replication studies. Based on the results of these analyses and our literature survey, we conclude that the current state-of-the-art in fully-automated misinformation detection has limited efficacy in detecting human-generated misinformation. We offer recommendations for evaluating applications of machine learning to trust and safety problems and recommend future directions for research. 

\end{abstract}

\section{INTRODUCTION}
%-------------------------------------------------------------------------------
\label{sec:intro}

Online services face a daunting task: There is an unceasing deluge of user-generated content, on the order of hundreds of thousands of posts per minute on popular social media platforms~\cite{dewan_2015}. Some of that content is false, hateful, harassing, extremist, or otherwise problematic. How can platforms reliably and proactively identify these ``trust and safety'' issues?

Machine learning has proved an attractive approach in the academic literature, leading to large bodies of scholarship on misinformation detection~\cite{alam_survey_2021}, toxic speech classification~\cite{andročec_2021}, and other core trust and safety challenges (e.g.,~\cite{lee_2020}). The conceptual appeal of machine learning is that it could address the massive scale of user-generated content on large platforms \textit{and} the capacity constraints of small platforms. Recent work claims impressive performance statistics: In the literature review that we conduct for this work, among publications that report performance metrics, about 70\% of papers report over 80\% accuracy on at least one detection task; some of these works report near-perfect performance \cite{baly_predicting_nodate, dhamani_using_2019}. 

In recent years, news items from major tech companies have tempered these expectations. In October of 2023, in a Bluesky post commenting on Twitter's user-driven Community Notes program, Twitter's former head of trust and safety stated that large-scale automated detection of misinformation remains a hard problem, and that no generalizable automated solutions are currently available \cite{roth_2023}. In January of 2025, Meta announced that it would terminate its third-party fact-checking program in favor of a Community Notes-like system of user-driven content moderation \cite{chan_meta_2025}. These disclosures accord with our observation that, in practice, trust and safety functions at online services remain heavily manual: driven by user reports and carried out by human moderators.
 
In this work, we investigate the disconnect between scholarship and practice in applications of machine learning to trust and safety problems. Our project is inspired by recent research that has identified shortcomings in machine learning applications for many problem domains, including information security~\cite{arp_ml_2022, Jacobs2022}. We use misinformation detection as a case study for trust and safety problems because the topic has recently generated a rich literature with diverse methods and claimed successes. Misinformation detection has substantive complexities that are common for trust and safety problems: linguistic and cultural nuance, sensitivity to context, and rapidly evolving circumstances. 

We seek to answer four discrete research questions, which collectively shed light on the research-practice gap in automated misinformation detection.

\begin{itemize}[nolistsep]
\item[\textcolor{blue}{RQ1.}] How well-suited are misinformation detection methods in the academic literature to the needs of online services, specifically social media platforms that host user-generated content? 

\item[\textcolor{blue}{RQ2.}] Are there identifiable missteps related to target selection, dataset curation, feature selection, and evaluations of method performance in ML-driven misinformation detection studies? 

\item[\textcolor{blue}{RQ3.}] How reproducible are published ML-driven misinformation detection methods?

\item[\textcolor{blue}{RQ4.}] How generalizable are published ML-driven misinformation detection methods to out-of-domain data (i.e., to data types and topics not present in training data)?
\end{itemize}

We address these research questions in three ways. First, we conduct a broad literature review and synthesis of the full paper corpus (248 papers), with a focus on detection targets and evaluation. We provide an in-depth review of a subset (87 papers) of the full paper set, with a focus on methods: dataset design, feature engineering, and model selection. Second, we attempt to obtain code and data for a subset of prior work. Third, we test several representative approaches for replication and generalizability. We arrive at the following results by applying these methods. 
\begin{enumerate}[noitemsep]
    \item Detection tasks in scholarship are often steps removed from the misinformation content moderation challenges that platforms face. Detection targets may be of limited consequence, or may be more readily and accurately identified through manual means.

    \item Methods frequently target \textit{proxies} for the presence of misleading content; these approaches are easy to evade. Datasets used in publications are often non-representative of real-world contexts. Model evaluation often lacks independence from training and rarely involves close emulation of a real-world deployment. 

    \item Data and code availability problems pervade the literature, inhibiting replication. Where these are available, we are generally able to replicate prior work. 

    \item Prior work has poor generalizability when classifying content beyond what was included in training data. 
    
\end{enumerate}

Through this work, we hope to underscore 1) the prevalence and severity of reproducibility and ML-driven method design issues in the existing misinformation detection literature, 2) the need for careful and preemptive evaluation of ML-driven methods at the point of problem formulation, and 3) the importance of method explainability and data accessibility. Based on these observations, we provide recommendations for future work that proposes ML models to address trust and safety concerns. \textbf{We contribute information taxonomies, our annotated corpus of 248 research papers, recent datasets, and frameworks for evaluating ML-driven content moderation tasks.}

\section{MOTIVATION}
\label{sec:motivation}

The detection of misinformation and ``influence operations'' (IOs) is a topic of convergent interest for security, social science, and AI/ML researchers. Methods developed in recent years for the detection of influence operations (e.g., astroturfing, coordinated misinformation campaigns) resemble techniques previously employed by security researchers for the detection of botnets, advanced persistent threats (APTs), and malware \cite{milajerdi2019holmes,king2006subvirt,stone2009botnet}. As such, we believe that the security community is uniquely well-positioned to shepherd the responsible development of misinformation detection methods. In addition, misinformation research stands to benefit a good deal from the formal structure of security research methodology: In a survey study from 2022 conducted by Mirza et al., human fact-checkers and journalists expressed interest in adopting rigorous threat modeling practices similar to those found in academic security literature \cite{mirza2023tactics}. 

\section{PRELIMINARIES}
\label{sec:preliminaries}

\textbf{Definitions.} In the absence of agreed-upon definitions of \textit{misinformation, disinformation, malinformation,} and \textit{influence operations}, we refrain from advancing singular definitions of the same\footnote{While certain taxonomies \cite{alam_survey_2021, zhou_survey_2020, oshikawa_survey_2018} provide relative definitions of these terms---distinguishing \textit{misinformation} from \textit{disinformation}, for instance, via the presence or absence of intent---we emphasize that, to the best of our knowledge, there is no robust invariant for identifying a statement as (mis)informative on the basis of \textit{semantics alone}.}. Instead, for each paper we review, we consider the paper authors' working definitions of these terms (if such definitions are provided) and evaluate model performance with respect to the definitions set forth \cite{wu_misinformation_2019,gelfert_fake_2018,haiden_definitional_2018,klein_fake_2017,tandoc_defining_2018}. We limit our analysis to text-based English-language misinformation.\footnote{We acknowledge that the nature of misinformation narratives and misinformation spread varies with language and geography.} We note that \textit{disinformation} is commonly used to refer to incorrect information written with the intent to deceive, while \textit{misinformation} refers to incorrect information in general; for the sake of completeness and concision, we refer to intentionally \textit{and} unintentionally false information as \textit{misinformation} for this work unless otherwise qualified. We occasionally use \textit{false rumors,} \textit{false news,} and \textit{false information} interchangeably, but avoid \textit{fake news}, a politically charged term \cite{habgood_coote_2018}. We use ``influence operations'' (IOs) or ``coordinated campaigns'' in our discussion of collaborative attempts to disseminate misinformation across networks \cite{jackson_2023}.

\textbf{Feasibility of detection.} Most of the work we review presupposes that automated detection of misinformation is possible at all. We note that this is, in itself, a strong assumption to make: Published work in psychology, linguistics, and philosophy of language has previously called into question the feasibility of using semantic and syntactic features to determine the veracity of text statements and the usefulness of binary true/false classifications of information, particularly in the absence of unified definitions of misinformation, disinformation, and information \cite{soe_2017,vrij_pitfalls_2008}. On the other hand, some researchers maintain that misinformative texts are characterized by indelible ``fingerprints'' that distinguish them from non-misinformative texts: for instance, emotional language and reduced lexical diversity \cite{amado_undeutsch_2015,carrasco_2022}. This debate motivated our discussion of non-textual feature sets for detection. 

\textbf{Taxonomies.} 
Through an iterative process of reading and inductive coding of our papers, we develop a taxonomy of five ``information scopes.'' As we read each paper in our corpus, we noted the \textit{operative unit} of misinformation detection for the classifier or method described---the smallest semantic or organizational unit of information that the method attempted to classify as true or false---and sorted each paper into emergent categories. Our final taxonomy comprises five information scopes: claims, news articles, social media accounts, networks, and websites. We consider these scopes \textit{in the context of} online services---for instance, news articles and websites whose links might be posted to a social media platform.\footnote{We label certain works as members of \textit{multiple} scopes when they make classification decisions about more than one type of detection unit: works targeting social media posts, for instance, are generally categorized as falling within (\circled{C} \textit{and} \circled{U}) or (\circled{C} \textit{and} \circled{N}) in~\Cref{app:full_coding}.} 

\begin{itemize}
\item[\circled{C}] \textbf{Claims}. The smallest semantic unit of fact or misinformation, comprising a subject, predicate, and object, at minimum. 
\item[\circled{A}] \textbf{Articles}. News-oriented writing of length 100 words or more.
\item[\circled{U}] \textbf{Users}. All data and metadata associated with a single user's account as defined by a social media platform (e.g., an account corresponding to a handle on Twitter/X; or a page or profile corresponding to one business, individual, or organization on Facebook). 
\item[\circled{N}] \textbf{Networks}. A set of users and interactions between these users as represented by a social graph. 
\item[\circled{W}] \textbf{Websites}. A news site, including its hosting infrastructure and text and image contents.
\end{itemize}

We describe errors in method design with respect to the step of the development pipeline in which they occur. We break this process down as follows: target selection, dataset curation, model choice, feature set selection and model evaluation. Development step definitions follow: 

\begin{itemize}
\item[\circled{1}] \hl{\textbf{Target selection.}} The \textit{stated} versus \textit{actual} objective(s) of the classification task that paper authors describe: for instance, detection of emotional valence of a text (angry, sad, happy); verification of semantic accuracy (true, false); or characterizing degree of virality (e.g., as measured by number of reshares on a post). 

\item[\circled{2}] \hlcyan{\textbf{Dataset curation.}} The source, size, and contents of datasets used for model training and testing. Provenance information includes temporal labels for 1) the date of the dataset's production and 2) the date of dataset access by paper authors. 

\item[\circled{3}] \hlred{\textbf{Model choice.}} The choice of ML model used by paper authors for their detection task, as well as their motivation for this choice. 

\item[\circled{4}] \hlgreen{\textbf{Feature selection.}} The choice of feature(s) that paper authors use to train their ML models, as well as their motivation for this choice. 

\item[\circled{5}] \hlorange{\textbf{Model evaluation.}} Paper authors' approach to benchmarking method performance after initial training. This includes 1) choice of test dataset; 2) performance statistics (e.g., ROC values, true and false positive rates); and 3) ecological validity of test cases in relation to proposed deployment setting. 
\end{itemize}

\textbf{Paper organization.} In \textbf{\Cref{sec:motivation}}, we situate this project within existing security literature and academic literature that critiques machine learning-driven methods. We motivate our choice of automated misinformation detection as a representative case study and highlight the particular relevance of this problem to the security community. In~\Cref{sec:preliminaries}, we present the information taxonomies developed from our inductive coding of papers. We discuss findings from our reading and coding of papers in~\textbf{\Cref{sec:systematization_of_literature}}. Our systematization of literature progresses along two axes: 1) from unimodal to increasingly multi-modal methods, and 2) in order of operations of method development. Specifically, we discuss issues pertaining to method-target fit, dataset curation and feature selection, model selection, and method evaluation (\textcolor{blue}{RQ1}, \textcolor{blue}{RQ2}). In~\textbf{\Cref{sec:rep_studies}}, we illustrate the issues identified in our literature review with a series of replication studies, with a focus on reproducibility and generalizability of results (\textcolor{blue}{RQ3}, \textcolor{blue}{RQ4}). In~\textbf{\Cref{sec:discussion}}, we conclude with 1) a discussion of findings from our literature review and replication studies and 2) recommendations for evaluating ML-driven interventions for trust and safety.

\section{Systematization of Literature}
\label{sec:systematization_of_literature}

\textbf{Paper selection.} To seed our corpus, we manually curated a selection of 23 highly-cited survey papers that provide comprehensive overviews of the state of automated misinformation detection at the time of writing \cite{khan_benchmark_2021,fallis_functional_2014,zhou_survey_2020,alam_survey_2021,oshikawa_survey_2018,conroy_automatic_2015,sharma_combating_2019,islam_deep_2020,ahmed_detecting_2021,shu_fake_nodate,habib_false_2019,zhang_fake_2018,de_oliveira_identifying_2021,shu_mining_2020,miro-llinares_misinformation_2021,chen_news_2015,fernandez_online_2018,martin_recent_nodate,shelke_source_2019,guo_future_2019,thorne_automated_2018,zubiaga_analysing_2016,rubin_deception_2015}. We relied on these papers to ground our own understanding of existing approaches to automated misinformation detection, and to identify detection methods that have been well-received by the research community. We searched for survey papers in Google Scholar with queries ``survey misinformation detection'' and ``survey fake news detection'' and collected the most-frequently cited papers within the past 10 years.\footnote{This time frame was naturally enforced by a lack of well-cited older publications, and was not fixed before we began our sampling process.} We then inspected each paper's reference section for related papers; we read the abstracts for these references in order to confirm relevance and fit. We supplemented this core corpus of highly-cited papers with publications surfaced by Google Scholar queries. We queried the following terms on Google Scholar: ``\texttt{misinformation detection} [{$x$}]'' and ``\texttt{automated fact checking} [$x$],'' where $x$ $\in \lbrace$claims, news articles, accounts, networks, websites, influence operations\footnote{We include ``influence operations'' as a separate search term because this term of art is a relatively new one---we group most of these works with our network-scoped methods.}$\rbrace$. We collected the 50 most highly-cited papers in the set resulting from the union of search results returned by both search queries for each $x$. To counter potential bias toward older publications, we collected papers with the highest citation rates \textit{per year}. We note that these search terms are deliberately over-inclusive; we manually review all potential works for relevance after the initial sampling step. After removal of out-of-scope works (see \textit{In- and out-of-scope work}) from this set of 250 papers, 219 eligible papers remained. To ensure that security-oriented approaches to detection were represented in our corpus, we conducted a separate snowball sampling search for work published in security venues: Using the same keywords listed in the previous subsection, we oversampled publications from four A*\ \cite{core} security research venues (USENIX Security, IEEE S\&P, NDSS, and ACM CCS). This process resulted in the addition of 29 works, most of which address the detection of accounts and networks that spread misinformative content (e.g., botnets and trolls). Our final corpus comprises 248 papers published between 2009 and 2024, inclusive.

\textbf{``Full'' and ``focus'' paper corpora.} We conduct our literature survey at two different levels of granularity. For all papers in the full corpus (248 works), we note detection targets and model evaluation (steps \circled{1} and \circled{5} in~\Cref{sec:preliminaries}). For a subset of these (87), we perform deep-coding of methods: we note dataset design choices, model selection, and feature sets (resp. steps \circled{2}, \circled{3}, and \circled{4} in~\Cref{sec:preliminaries}).\footnote{We take inspiration from \cite{wei_sociodem_2024, scheffler_e2ee_2023, warford_ux_2021}, who perform a similar deep-coding of a subset of their whole-paper corpus.} A partial summary of this deep-coding is available in~\Cref{app:partial_coding}. Our motivation for developing this focus set is practical: Many of the works we review do not include in-depth discussions of model choice, weights, and datasets. As such, each per-scope discussion is book-ended by overviews of the full paper corpus, with focused analyses of method design details in between. 

\begin{table}[ht!]
\small
\centering
\begin{tcolorbox}[
    colframe=black,      % Border color
    colback=white,       % Background color
    arc=5mm,             % Rounded corner radius
    boxrule=0.3mm,       % Border thickness
    width=8.5cm,    % Full width of the page
    % % enhanced,
    % halign=center        % Center align the content
]
\renewcommand{\arraystretch}{1.5} % Adjust row height
\setlength{\tabcolsep}{10pt}      % Adjust column spacing
\begin{tabular}{p{7cm}}
\textbf{\hl{Targets}} \\
\circled{C} Claims: \textit{i.} Content-based detection via distance calculations on semantic embeddings; \textit{ii.} Content-based detection via search on knowledge graph topology; \textit{iii.}~``Checkability'' or ``checkworthiness.'' \\ 
\circled{A} Articles: \textit{i.} Syntactic and stylometric signals, including genre and sentiment; \textit{ii.} Topic-aware detection of stance and relevance to known rumoring topics. \\
\circled{U} Users: \textit{i.} Account metadata (bios, images, account age); \textit{ii.} Single account behaviors (comments on posts, published posts). \\
\circled{N} Networks: \textit{i.} Propagation patterns across social graphs; \textit{ii.} Timestamped records of user-user interactions. \\ 
\circled{W} Websites: \textit{i.} Text and URL/domain semantics; \textit{ii.}~Site visitor demographics; \textit{iii.} Suspicious UI elements; \textit{iv.} Hosting infrastructure (DNS certificate, site age). 
 \\ 
 \midrule 
 \textbf{\hlcyan{Datasets}} \\
\textit{i.} Dataset age;
\textit{ii.} Evidence of leakage (temporal leakage, feature leakage?);
\textit{iii.} Data dependencies (author, source, style); 
\textit{iv.}~Availability of information required to reproduce or reconstruct similar datasets (was non-public information required to produce ground-truth training sets?); \textit{v.} Availability of original data. \\
 \midrule
\textbf{\hlred{Models}}  \\       
\textit{i.} Distance calculations on semantic embeddings; \textit{ii.}~``Traditional'' ML (SVM, RF, DT); \textit{iii.} Deep learning (CNN, LSTM, GRU); \textit{iv.} Graph cut algorithms; \textit{v.}~Stacked ensemble classifiers; \textit{vi.} Graph clustering algorithms. \\
\midrule
\textbf{\hlgreen{Features}}  \\ 
\textit{i.} Textual; \textit{ii.} Network-based; \textit{iii.} Author-, user- or source-based; \textit{iv.} Infrastructural. \\ 
\midrule
\textbf{\hlorange{Evaluation}} \\
\textit{i.} Testing in real time; \textit{ii.} Generalizability of approach; \textit{iii.}~Evasion-resistance; \textit{iv.} Robustness to distributional shifts in training or test data; \textit{v.} False-positive/false-negative rates. \\
\end{tabular}
\end{tcolorbox}
\caption{Taxonomy of targets, models, features, and evaluation codes. Alphanumeric abbreviations (e.g., ``\textbf{C.i}'') are referenced throughout this work.}
\label{tab:taxonomy}
\end{table}

\textbf{Corpus curation and coding.} As a first-pass relevancy check, we used automated keyword matching to confirm that all collected papers did, in fact, address misinformation and automated fact-checking methods; we read the abstracts of papers surfaced by this check to confirm relevance and fit. We then annotated research papers that passed this check in accordance with our codebook (see SI). One reader made three separate coding passes over the corpus, varying the objectives of her annotation on each pass: In the first pass, she specifically sought to identify taxonomies for classification and detection; on the second, she noted actual versus stated detection targets and approaches to evaluation; on the final pass, she noted method design approaches and errors. Two coders read and independently coded a random subset (30 papers) of the full corpus. Fleiss’s kappa for this subset was >~0.80, indicating strong agreement.

\textbf{In- and out-of-scope work.} We consider a number of security-oriented approaches to misinformation detection in this work. Sybil and botnet detection methods that specifically target influence operations online are categorized as account- and network-scoped detection methods in our corpus. \textit{Commercial approaches} to misinformation detection are in-scope for this project. In view of data accessibility issues, however, we are unable to provide an in-depth analysis of commercial methodologies in our main literature review, and instead include a market survey of commercial fact-checking providers in~\Cref{sec:commercial_llm_checking}.
(In general, commercial vendors do not make code and training data publicly available.) \textit{LLM-powered detection} is in-scope for this work. Accessibility to code bases for LLMs is similarly limited (this has been discussed at length in the popular press \cite{mollman_2023, wong_2024}), and disallows testing and evaluation by researchers. As such, while we do briefly discuss LLM-powered detection approaches in our supplement, we defer a more extensive discussion to future work.

\textbf{Notation.} In these sections, we denote information scope and pipeline steps with circled icons (\circled{A}, \circled{1}) and subtaxa within those categories with lowercase bolded Roman numerals (\textbf{iv}). We note that, while \textit{targets} are generally unique to each information scope (and are referenced by an alphanumeric label (e.g., \hl{\textbf{(A.i)}}), critiques of steps \circled{2}-\circled{5} are \textit{not} scope-specific, and are always denoted by Arabic-Roman numeral pairs (e.g., \hlcyan{\textbf{(2.i)}}). We highlight each subtaxon in-text, with a different color corresponding to each development step. (The full taxonomy is in~\Cref{tab:taxonomy}.) A summary of findings for each scope is available in the starred \textbf{**Takeaways} summary at the end of each subsection.

\circled{2} \textbf{Datasets.} Claim-scoped papers in our corpus that propose to perform fact verification \hlcyan{\textbf{(2.i)}} \textit{rely on outdated existing datasets} of labeled statements in order to establish ground truth. LIAR \cite{wang_liar_2017}\footnote{We note that the LIAR dataset is actually a collection of 27.8K labeled PolitiFact statements---so, in a sense, PolitiFact is the dominant data source.} and PolitiFact \cite{politifact} were the most popular datasets among claim-scoped works, and, taken together, were used by approximately half of all papers within this scope \cite{wang_learning_2019, soprano_many_2021}. LIAR is a static political news dataset that was published in 2017; on average, papers that cite LIAR were published two years after LIAR's release. Topic detection and word frequency models trained on LIAR are likely ineffective in contemporary fact-checking contexts, and for non-political subject matter \cite{debnath_hierarchical_2020, rasool_multi-label_2019}.\footnote{Choice of ground truth site or labeled dataset can significantly influence the outcome of analysis: Bozarth et al. found that perceived prevalence of misinformation in a corpus of 2016 election news varied from 2\% to 40\%, depending on choice of ground-truth reference website \cite{bozarth-truth-2020}.} Additionally, misinformation taxonomies across reference sites are inconsistent: PolitiFact employs a six-point labeling scale (pants-on-fire; false; barely-true; half-true; mostly-true; true), FEVER employs a three-point scale (supported; refuted; notenoughinfo), and GossipCop employs an eleven-point scale (ratings from 0 to 10). This divergence is likely a symptom of definitional issues (\Cref{sec:preliminaries}). 

\circled{3} \textbf{Model selection.} Model choice follows target choice for claim-scoped methods: for methods that pre-construct knowledge graphs or other reference databases, models perform some form of \hlred{(\textbf{3.i})} \textit{shortest-path search} on the KG topology \cite{shiralkar_finding_2017, ciampaglia-wiki-2015}, and approximate logical inference via transitive closure on graph edges. For methods that perform information retrieval at query time---e.g., to match corroborating sources to a claim to be checked---model training generally follows conversion of the text statement to a bag-of-words or TF-IDF embedding; choice of model is highly variable \cite{bhutani_fake_2019, shu_defend_2019}, and does not appear to be predictive of performance. For methods that perform detection of misinformative posts on social media, \hlred{(\textbf{3.v})} \textit{stacked ensemble classifiers} are a common approach to incorporating multiple feature modalities.

% \section{Paper-by-feature summary}
\begin{table*}[ht!]
\small
\centering
\resizebox{!}{1.55in}{
    \begin{tabular}{lc|c|c|c|cccc|r}
    \multicolumn{2}{c|}{\textbf{\ul{Paper}}} &
    \multicolumn{1}{c|}{\circled{1} \textbf{\ul{Target}}} &
    \multicolumn{1}{c|}{\circled{2} \textbf{\ul{Dataset}}} & \multicolumn{1}{c|}{\circled{3} \textbf{\ul{Model}}} & \multicolumn{4}{c|}{\circled{4} \textbf{\ul{Features}}} & \multicolumn{1}{c}{\circled{5} \textbf{\ul{Performance}}} \\
    \textbf{Work} & \textbf{Scope} &  &  &  & \begin{sideways}{\textbf{Textual}}\end{sideways} & \begin{sideways}{\textbf{Network \space}}\end{sideways} & \begin{sideways}{\textbf{Author}}\end{sideways} & \begin{sideways}{\textbf{Infra.}}\end{sideways} &  {\textbf{Accuracy/AUROC}} \\
    \midrule
1. Ajao et al. \cite{ajao_sentiment_2019} & \circled{C} \circled{A} & Sentiment \textbf{(A.i)} & PHEME \cite{pheme_2018} & LSTM, DT, RF, \underline{SVM} & \fullcirc &  & & & 0.86 (Acc.) \\

2. Abulldah-All-Tanvir et al. \cite{abdullah-all-tanvir_detecting_2019} & \circled{C} \circled{A} \circled{N} & Content \textbf{(C.i)} & Twitter (API) & NB, RNN, LSTM, \underline{SVM}, Logit & \fullcirc & \fullcirc & & & 0.89 (Acc.) \\

3. Bhutani et al. \cite{bhutani_fake_2019} & \circled{C} \circled{A} & Content \textbf{(C.i)}; sentiment \textbf{(A.i)} & Twitter (API), PolitiFact \cite{politifact} & Naive Bayes, \underline{RF} & \fullcirc & & & & 0.60 (AUC) \\

4. Bozarth et al. \cite{bozarth-truth-2020}  & \circled{C} & Contents \textbf{(C.i)} & PolitiFact \cite{politifact}, Daily Dot, Zimdars, MBFC & LDA & \fullcirc & & & & n/a \\

5. Ciampaglia et al. \cite{ciampaglia-wiki-2015} & \circled{C} \circled{N} & Shortest path search \textbf{(C.ii)} & DBpedia & kNN, RF & \fullcirc & & & & 0.97 (AUC) \\

\midrule

%%%%% end claims, begin articles %%%%%%%%%%

1. Afroz et al. \cite{afroz_detecting_2012} & \circled{A} & Content \textbf{(C.i)}; syntax \textbf{(A.i)}  & Brennan-Greenstadt & SVM, J48 Decision Trees & \fullcirc &  &  &  & 0.97 (F1) \\

2. Ahmed et al. \cite{ahmed_detecting_2018} & \circled{A}& Syntax \textbf{(A.i)} & Twitter, Kaggle, Horne and Adali \cite{horne_this_2017} & SVM & \fullcirc & & & & 0.92 (Acc.) \\

3. Bourgonje et al. \cite{bourgonje_clickbait_2017} & \circled{A} & Stance (\textbf{A.ii)} & Fake News Challenge Data & Logit & \fullcirc & & & & 0.90 (Acc.) \\

4. Brasoveanu et al. \cite{brasoveanu_semantic_2019} & \circled{A} \circled{C} & Sentiment (\textbf{A.i}); keywords \textbf{(A.ii)} & LIAR\cite{wang_liar_2017} & CNN, LSTM, \underline{CN} & \fullcirc & \fullcirc & & & 0.64 (Acc.) \\

5. Della Vedova et al. \cite{della_vedova_automatic_2018} & \circled{A} \circled{N} & Content \textbf{(C.i)}; virality \textbf{(N.iv)} &  FakeNewsNet, Buzzfeed & Logit & \fullcirc & \fullcirc & & & 0.82 (Acc.) \\

\midrule

%%%%% end articles, begin users %%%%%%%%%%

1. Cao et al. \cite{cao_aiding_2012} & \circled{U} \circled{N} & Acct. cred. \textbf{(U.i)}; prop. \textbf{(N.i)} & Tuenti social network & Louvain clustering &  & \fullcirc & \fullcirc & & 0.90+ (TP) \\

2. Danezis et al. \cite{danezis2009sybilinfer} & \circled{U} \circled{N} & Acct. cred. \textbf{(U.i)}; prop. \textbf{(N.i)} & LiveJournal data & Bayesian inf. & & \fullcirc & \fullcirc & & n/a*  \\

3. Ezzeddine et al. \cite{ezzeddine2023exposing} & \circled{U} & Acct. behaviors \textbf{(U.ii)} & DATA & LSTM & & \fullcirc & \fullcirc & & 0.91 (AUC) \\

4. Hamdi et al. \cite{hamdi_hybrid_2020} & \circled{U} \circled{N} & Account metadata \textbf{(U.i)}; prop. \textbf{(N.i)} & CREDBANK & LDA, Bayes, Logit, SVM & & \fullcirc & \fullcirc & & 0.99 (AUC) \\

5. Helmstetter et al. \cite{helmstetter-supervised-2018} & \circled{U} \circled{N} \circled{A} & Acct metadata \textbf{(U.i)}; post sharing data \textbf{(U.ii)} & Public site cred. lists & SVM, NB, DT, RF & \fullcirc & \fullcirc & \fullcirc & & 0.936 (F1) \\

\midrule

%%%%% end users, begin networks %%%%%%%%%%

1. Alizadeh et al. \cite{alizadeh2020content} & \circled{N} \circled{A} \circled{U} & Propagation \textbf{(N.i)}; syntax \textbf{(A.i)}; acct metadata \textbf{(U.i)} & Twitter (API), Reddit IRA troll list & RF & \fullcirc & \fullcirc & \fullcirc & & 0.70+ (F1) \\

2. Antoniadis et al. \cite{antoniadis_model_2015} & \circled{U} \circled{A}  & Acct metadata \textbf{(U.i)}; syntax \textbf{(A.i)} & Hurricane Sandy tweet dataset & J48, \underline{RF}, KNN, Bayes & \fullcirc & & \fullcirc & & 0.79 (Avg. Prec.) \\

3. Assenmacher et al. \cite{assenmacher2020two} & \circled{N} \circled{A} & Propagation \textbf{(N.i)}; topic det. \textbf{(A.ii)} & Twitter (API) & Clustering & \fullcirc & \fullcirc & & & not reported \\

4. Buntain et al. \cite{buntain_automatically_2017}  & \circled{N} \circled{U} \circled{A} & Time \textbf{(N.ii)}; acct metadata \textbf{(U.i)}; sentiment \textbf{(A.i)} & CREDBANK, Buzzfeed & RF & \fullcirc & \fullcirc & \fullcirc & & 0.65 (Acc.) \\

5. Castillo et al. \cite{castillo_information_2011} & \circled{U} \circled{A} & Syntax \textbf{(A.i)}; user behavior \textbf{(U.ii)} & Twitter Monitor events & SVM, DT & \fullcirc & \fullcirc & \fullcirc & & 0.874 (P) \\
\midrule

%%%%% end networks, begin websites %%%%%%%%%%

1. Asr et al. \cite{torabi_asr_data_2018} & \circled{W} \circled{A} & Source rep. \textbf{(W.i)}; Syntax \textbf{(A.i)} & BuzzfeedUSE, Snopes, Rashkin, Rubin & CNN, \underline{SVM}, NB & \fullcirc & \fullcirc &  & \fullcirc & not reported \\
2. Baly et al. \cite{baly_predicting_nodate} & \circled{W} \circled{A} \circled{N} & Source rep. \textbf{(W.i)}; Site infra. \textbf{(W.ii)}; \textbf{(A.i)} & MediaBiasFactCheck\cite{mediabiasfactheck} &  SVM & \fullcirc & & & \fullcirc & 0.66 (Acc.) \\ 
3. Baly et al. \cite{baly-etal-2020-written} & \circled{W} \circled{A} \circled{N} & Source rep. \textbf{(W.i)}; Site infra. \textbf{(W.ii)}; \textbf{(A.i)} & MediaBiasFactCheck\cite{mediabiasfactheck} & SVM & \fullcirc & & & \fullcirc & 0.7152 (Acc.) \\ 
4. Castelo et al. \cite{castelo_agnostic_2019} & \circled{W} \circled{A} & Site infra. \textbf{(W.ii)}; syntax \textbf{(A.i)} & Celebrity, US-Election2016 & \underline{SVM}, kNN, RF & \fullcirc & & & \fullcirc & 0.86 (Acc.) \\
5. Chen et al. \cite{chen_proactive_2020} & \circled{W} \circled{A} & Hosting infra. (URL) \textbf{(W.ii)}; syntax \textbf{(A.i)} & PoliticalFakeNews & Clustering & \fullcirc & & & \fullcirc & 0.97 (AUC) \\
    \end{tabular}
}
\caption{\textbf{Focus corpus by scope and target.} Coding of 25 papers in our focus set, sorted by information scope. Codes for the full focus set appear in~\Cref{app:full_coding}. Values in parentheses in ``Target'' field correspond to highlighted subcategories presented in~\Cref{sec:systematization_of_literature} (e.g., ``C.i'' denotes target (i) in ``Claims,'' ~\Cref{sec:claims}). If authors present evaluation results for multiple models, we \underline{underline} the most performant model and record its corresponding performance score.}
\label{app:partial_coding}
\end{table*}

\subsection{Claims} 
\label{sec:claims}
% For purposes of discussion, we define a single ``claim'' here as the smallest semantic unit of fact or misinformation. 
About 70\% of paper authors within this scope cite the 2016 U.S. presidential election as motivation for the development of their methods \cite{debnath_hierarchical_2020, hassan_data_2014, katsaros_which_2019}; about 30\% cite COVID-19 misinformation \cite{kou_hc-covid_2022, roitero_can_2021, glockner_missing_2022}. All works mention the speed and volume of \textit{social media} misinformation in particular \cite{kartal_too_2020}. 30\% cite the relatively slow pace of manual fact-checking \cite{ciampaglia-wiki-2015,hassan_data_2014, kartal_too_2020}---and the need for faster, automated approaches---as motivation. Claim-scoped papers form 15\% of our corpus.

\circled{1}
\textbf{Detection targets.} Across all information scopes, we find that claim-scoped detection methods are most consistent in their attempts to verify semantic contents of text statements. This is done in the following ways: \hl{\textbf{(C.i)}} \textit{distance calculations on semantic embeddings} to perform textual entailment or stance detection \cite{saikh_novel_2019, borges_combining_2018}; and \hl{\textbf{(C.ii)}} \textit{search on a knowledge graph topology} \cite{ciampaglia-wiki-2015, shiralkar_finding_2017} to determine if these reference sources corroborate or refute the claim to be checked. A small class of approaches explicitly detect \hl{\textbf{(C.iii)}} ``\textit{checkworthiness},''~\footnote{We include these works in our corpus because they \textit{do} take topic and source credibility into consideration in the process of ranking checkability.} and are intended to surface checkable statements to human fact-checkers for manual verification \cite{kartal_too_2020, shiralkar_finding_2017, hassan_data_2014}. 

Targets such as author credibility and language cues are \textit{proxy targets} for the presence of misinformation: signals which may not be sufficient for determining text veracity in isolation, but which are indicative of (lack of) veracity by association with an external heuristic. About 60\% of papers within the full corpus at this scope employ non-semantic targets, including the \hl{\textbf{(A.i)}} \textit{syntactic and/or stylometric qualities of text} \cite{jadhav_fake_2019}, \hl{\textbf{(U.i)}} \textit{source reputation} \cite{rasool_multi-label_2019}, or \hl{\textbf{(U.ii)}} \textit{contextual indicators}, such as commenter responses \cite{tian_early_2020}, to classify social media posts.  

\circled{4} \textbf{Feature selection.} At the level of single claims, semantic feature analysis is limited to the \hlgreen{\textbf{(4.i)}} \textit{identification of structured statements} as a precursor to knowledge graph (KG) construction. These claims take the form of subject-predicate-object (SPO) statements (e.g., ``I like pie'')~\cite{ciampaglia-wiki-2015, magdy-stat-2010, shiralkar_finding_2017}. Detection versatility is determined by the size of the source dataset and the granularity of the relationships encoded by graph edges \cite{shiralkar_finding_2017}. Supervised methods that detect linguistic cues employ \hlgreen{\textbf{(4.i)}} \textit{hand-crafted word or topic lists} \cite{kartal_too_2020}). Authors employing supervised methods claim that their approach permits highly customized targeting of specific rumoring narratives \cite{kartal_too_2020}, though this also assumes that the method developer has prior knowledge of the contents of test data; authors employing unsupervised methods claim that their approaches detect contextual and language features that cannot be easily extracted by common features such as word frequency or sentiment \cite{wang_learning_2019}. Methods detecting social media data consider \hlgreen{\textbf{(4.ii)}} \textit{network} and \hlgreen{\textbf{(4.iii)}} \textit{user} interaction features (post likes, shares, comments) \cite{paudel_lambretta_2023, gupta_tweetcred_2015, cui_same_2019}.

\circled{5} \textbf{Evaluation.}
Though a majority of claim-scoped methods cite the speed of social media misinformation as motivation, \hlorange{\textbf{(5.i)}} \textit{only one method within this scope reported results from testing in real time} \cite{hassan_toward_2017}. KG-based methods are \hlorange{\textbf{(5.ii)}} \textit{non-generalizable by design}: the approaches we survey require structured inputs for graph construction, and rely on published datasets for ground truth \cite{ciampaglia-wiki-2015,shiralkar_finding_2017}. Though this approach permits semantic verification of statements, it is difficult to perform iterative updates to source databases in real-time, particularly in the types of online settings where claim-checking might be most usefully deployed (e.g., during breaking news events, where no source of ground truth is immediately available). We observe that a majority of works at this scope do not test on novel or out-of-domain data; we discuss overfitting issues in greater detail in the next section. 

% \begin{tcolorbox}
\textbf{**Takeaways}: \circled{1} The efficacy of claim-scoped methods is completely determined by the depth and breadth of coverage conferred by a reference database. \circled{2} Datasets are frequently out-of-date, and taxonomies are inconsistent. \circled{3} Though some inference is possible via transitive closure on knowledge graph edges, this capability is, in general, limited. \circled{4} Knowledge-graph-based methods require structured inputs, which might not be readily available in a breaking news setting, or in scenarios where ground truth references are not yet available. \circled{5} Few authors test in real-time, despite citing the slow pace of manual fact-checking as motivation.
% \end{tcolorbox}

% \textbf{Recommendations}: Existing methods that rely upon third-party references should regularly update their graph relationships to reflect updates to the reference database.

%------------------------------------------------------------%

%----------------------- articles -----------------------%

\subsection{Articles} 
\label{sec:articles}
We consider all news-oriented writing of length 100 words or greater to fall within this scope. 25\% of papers within this scope cite growing distrust of mainstream news media outlets as motivation for their methods, which promise to deliver fast labeling of news stories that appear on social media \cite{rubin_deception_2015}. Text-based credibility classifiers have been shown to have limited efficacy, however: while unsupervised approaches can identify \textit{bias} with high accuracy, this performance degrades in misinformation and credibility classification tasks \cite{fairbanks2018credibility, potthast-etal-2018-stylometric}. This scope comprises 24\% of our paper corpus. 

\circled{1} \textbf{Detection targets.} In contrast to single claims, full-length news articles have sufficient text contents to make semantic verification difficult---and certain off-the-shelf NLP approaches practicable. These approaches are distinct from direct claim verification and qualify as proxy detection methods: For instance, Bhutani et al. associate strongly negative sentiment with the presence of false information \cite{bhutani_fake_2019}, and Horne et al. find that satire and misinformation share stylistic similarities \cite{horne_this_2017}. All article-scoped methods target proxy signals, and adopt at least one of the following three approaches to detection: \hl{\textbf{(A.i)}} \textit{NLP analysis of article contents to identify language features particular to writing styles heuristically associated with misinformation} (genre detection, sentiment analysis) \cite{mihalcea-strapparava-2009-lies}; and \hl{\textbf{(A.ii)}} \textit{analysis of article contents and headlines to identify potentially clickbait-y titles or discussion related to known misinformation narratives} (topic detection) \cite{zhang_fake_2018, kartal_too_2020}. 
Respectively, these approaches 1) simultaneously assume and detect a heuristic (e.g., strong emotion indicating the presence of misinformation); and 2) assume prior knowledge of rumoring topics.

\circled{2} \textbf{Datasets.} While well-annotated, current datasets are in short supply across all information scopes, this deficit is particularly glaring at the article scope. This is due in large part to definitional ambiguities that prevent fine-grained labeling of longer texts for classifier training. As a result, 44\% of paper authors within this scope \hlcyan{\textbf{(2.i)}} \textit{use public datasets released years prior to the start of their research} \cite{brasoveanu_semantic_2019,wang_liar_2017,horne_this_2017,shu_beyond_2019}. These datasets (LIAR \cite{wang_liar_2017}, Buzzfeed-Webis \cite{potthast-etal-2018-stylometric}, and PolitiFact \cite{politifact}) include news links, speaker credibility scores, and other metadata that \hlcyan{\textbf{(2.ii)}} \textit{constitute serious sources of leakage} for methods that use contextual features to infer true/false labels. The remaining 56\% of authors curate their own article corpora by asking crowdworkers to generate misinformative text \cite{perez-rosas_automatic_2017}, selectively editing true news articles (e.g., via verb inversion or noun replacement) \cite{mihalcea_lie_2009}, or compiling articles from authoritative news sources and known satire sites \cite{rubin_fake_2016}. These data curation techniques \hlcyan{\textbf{(2.iii)}} \textit{introduce additional dependencies and shortcuts} to textual datasets for which such variables are already difficult to control. Style \cite{potthast-etal-2018-stylometric} and genre \cite{hosseinimotlagh_unsupervised_nodate}, for instance, are emergent qualities of writing that cannot be easily marginalized out of a text embedding. 

\circled{3} \textbf{Model selection.} Among papers that report testing with multiple models---including \hlred{(\textbf{3.ii})} \textit{classical ML models} \cite{ahmed_detecting_2018}, \hlred{(\textbf{3.iii})} \textit{unsupervised NN models} \cite{brasoveanu_semantic_2019}, and \hlred{(\textbf{3.v})} \textit{stacked ensemble classifiers} \cite{ruchansky_csi_2017}---there is no clear correlation between model choice and actual performance. We note that, in instances where authors test on two- \textit{and} multi (i.e., $>$ 2)-way classification tasks, performance declines sharply in the latter case \cite{afroz_detecting_2012}. In those instances, reported performance scores are for two-way tasks. For this reason, as well, classical ML models (logit, SVM) oftentimes \textit{appear} to be most performant.

\circled{4} \textbf{Feature selection.} Among supervised methods that disclose their feature sets, we find that \hlgreen{\textbf{(4.i)}} \textit{word frequency, sentiment, and genre} were among the most commonly used features, and were collectively employed by 80\% of works within this scope; these features can also be sources of dependency-induced noise. It is difficult to quantify the impact of dependencies related to voice, house style, and source on classifier performance, particularly in the case of unsupervised learning methods, which comprise 59\% of methods at this scope. We evaluate an unsupervised learning method (and consider its performance in light of possible style-related dependencies) in our replication analysis of Nasir et al. (\Cref{sec:article_rep}) \cite{nasir_fake_2021}.

\circled{5} \textbf{Evaluation.}
Misinformation detection methods scoped to full texts risk overfitting to single topics: 42\% of authors select \hlorange{\textbf{(5.ii)}}  \textit{one or more narratives of interest} (e.g., the 2016 presidential election, the Boston Marathon bombing), train a classifier on these topics, then test this classifier on a different set of texts that discuss the \textit{same} topic \cite{noauthor_rumors_2014,jin_detection_2017,nasir_fake_2021}. This approach, while valid for evaluating classifier performance on closed datasets, lacks ecological validity for the use cases that authors claim that their methods will address: Rapid topic identification and high-quality annotation of relevant articles are generally unavailable in breaking news scenarios on social media platforms \cite{shu2020detecting}. 60\% of authors at this scope compare the performance of their detection method to other published approaches or ML models, but \hlorange{\textbf{(5.ii)}} \textit{neglect to test on novel datasets}. These methods do well when tested on in-domain texts, and in comparison to a selection of older ML models; many report accuracy well above 80\% \cite{zhang_detecting_2019,silva_embracing_2021}. Only one paper within the article scope tested in an adversarial setting: Its authors found that, while stylometry-based misinformation detection had an accuracy rate greater than 80\% on routine tasks, this score dropped to about 50\% in adversarial cases \cite{afroz_detecting_2012}. We demonstrate such a dropoff in \Cref{sec:article_rep}.

\textbf{**Takeaways}: \circled{1} In the absence of semantic definitions of misinformation, \textit{proxy} detection targets are common but easy to evade. \circled{2} Well-labeled datasets are rare; those datasets that are available are at least several years old at time of writing. \circled{3} Unsupervised methods show marginal improvements over classical ML models in some cases; it is unclear if these improvements are 1) significant or 2) sustainable across different datasets. \circled{4} Detection methods at the article scope are uniquely susceptible to text-based dependencies that are difficult to control for. \circled{5} Inflated performance scores can often be attributed to testing on same-topic news articles. 
% \end{tcolorbox}

\subsection{Users} 
\label{sec:accounts}
Evidence of foreign interference during the 2016 U.S. presidential election triggered a resurgent interest in malicious account detection \cite{ezzeddine2023exposing,saeed2022trollmagnifier}. As such, 90\% of security- and social-science-oriented works that we include within this scope (a dozen papers) explicitly discuss Russian trolls or other influence operations conducted by nation state actors and train classifiers on published lists of such accounts \cite{saeed2022trollmagnifier, zannettou2019disinformation, mirza2023tactics, lukito2020coordinating, keller2020political}. Account-scoped papers formed approximately 15\% of our corpus (40 papers).

\circled{1} \textbf{Detection targets.} Papers within this scope target source reputation, and \hl{\textbf{(U.i)}} \textit{inspect account metadata}, such bios, account age, and profile images; or distinguish suspicious accounts by \hl{\textbf{(U.ii)}} \textit{a single user's social behaviors}, such as their comments on posts (\textbf{n.b.} this target is distinct from \hl{\textbf{(N.i)}}). The security literature we review discusses trolls and bots deployed for astroturfing, misinformation campaigns, and IOs \cite{saeed2022trollmagnifier, cresci_bots_2020, chu2012detecting, zhang2016rise, alizadeh2020content}. In the absence of rigorous definitions of these account types, however, actual detection targets are tautological: A troll or bot is an account that exhibits troll- or bot-like behavior, or that interacts with confirmed troll or bot accounts \cite{saeed2022trollmagnifier, cresci_bots_2020}.

\circled{2} \textbf{Datasets.} All troll and bot account detection works we reviewed relied on published lists of ``known'' troll accounts for model training, but \hlcyan{\textbf{(2.iv)}} \textit{neglected to mention the heavily manual investigation required to produce these original lists }\cite{silverman_timberg_kao_merrill_2022, linvill2020troll}. Researchers who compiled some of these account lists, including a set of several hundred Twitter accounts with possible links to a known Russian troll farm (the Internet Research Agency, or IRA) manually examined suspicious accounts and tweet contents in order to produce detailed account and content taxonomies; notably, these classifications required external intelligence about account activity that was not published alongside account lists \cite{linvill2020troll}. Two well-cited troll lists compiled by the U.S. government, comprising thousands of suspicious Twitter and Facebook accounts, were curated using proprietary non-public information \cite{exhibit_b}.

\circled{3} \textbf{Model selection.} Detection proceeds via classifier training on a list of ``known'' suspicious accounts and application of this classifier to a dataset of novel accounts \cite{saeed2022trollmagnifier,lukito2020coordinating}. We note that, regardless of model choice, if classifier training data \textit{and} feature selection reflect a heuristic about suspicious behavior, the resulting classifier will simply learn this heuristic: The methods we review can be used to detect accounts whose behaviors conform to heuristic assumptions, but cannot be used to surface novel malicious behaviors, and are not resistant to attacks or evasion \cite{wang2014man}; we explore this further in our replication analysis of Saeed et al. (\Cref{sec:trollmagnifier}) \cite{saeed2022trollmagnifier}. A subset of methods at this scope and the network scope formulate detection as a \hlred{(\textbf{3.iv})} \textit{graph cut or influence maximization problem}, and describe approaches to identifying optimal cuts for isolating suspicious accounts \cite{zhang_misinformation_2016, cao_aiding_2012}.

\circled{4} \textbf{Feature selection.} Methods that define suspicious accounts by intrinsic \textit{properties} of these accounts (e.g., user handles and profile images) target the \hlgreen{\textbf{(4.iii)}} \textit{semantics of this account metadata}, and detect evidence of manipulation in (e.g.) image metadata and bios; or text outputs, such as posts and links \cite{helmstetter-supervised-2018, sansonetti_unreliable_2020}. Methods that define suspicious accounts by account \textit{activity} target \hlgreen{\textbf{(4.ii)}} \textit{networked behaviors}, such as liking and resharing statistics \cite{pacheco_uncovering_2020}. Feature sets for some methods in the first category include demographic data for users, such as inferred political party affiliation or race \cite{ribeiro_media_2018} (these \textit{n}th order assumptions are dangerous to make \cite{freelon_black_2020}; see our discussion about proxy signals, in~\Cref{sec:articles}).

\circled{5} \textbf{Evaluation.} We observe accuracy scores above 80\% for \hlorange{\textbf{(5.ii)}} \textit{confirmatory detection of like accounts} for all methods that reference a seed list of known trolls \cite{saeed2022trollmagnifier,addawood_linguistic_2019,ezzeddine2023exposing}; de novo detection of behaviors not represented within training data is not possible, by the self-admission of 10\% of authors within this scope \cite{saeed2022trollmagnifier, stone2009botnet}. As we discuss further in the next subsection (\textit{Networks}), the increasingly hybrid nature of IOs requires more nuanced taxonomies for classification: \textit{extent} of coordination, rather than \textit{existence}, might be a more appropriate measure of possible manipulation. \footnote{In a blog post on bot detection from 2021, Twitter ``debunked'' four common heuristics commonly used to identify bot accounts, including several detected by methods in our corpus, and described a ``forensic team of investigators'' who manually verify bot-ness of suspect accounts \cite{Thread_2021}.} Some authors of bot detection methods acknowledge that their approaches are \hlorange{\textbf{(5.iii)}} \textit{trivially easy to evade} if account holders 1) avoid interacting with known suspicious accounts or 2) vary their account identity and posting semantics \cite{danezis2009sybilinfer,cresci_bots_2020}.

\textbf{**Takeaways}: \circled{1} Targets are frequently tautological. \circled{2} Hand-annotated training datasets are the result of intensive fact-finding on the part of human researchers, and often require information that is not publicly available. \circled{3} Classifiers can only detect accounts resembling those in seed lists. \circled{4} Features that attempt to infer user credibility from demographic information risk reinforcing existing biases.  
\circled{5} Current methods cannot detect novel malicious behaviors.

\subsection{Networks} 
\label{sec:networks}
Within the security literature, a growing awareness of hybrid networks, which employ a combination of automated and manual approaches to disseminate content, has encouraged a turn toward \textit{network-based} bot detection methods, and away from detection of individual accounts \cite{cresci_bots_2020, grimme2018changing}. We observe a parallel turn toward network-based methods in AI, ML, and NLP venues as a result of growing recognition of overfitting and generalizability issues in purely text-based detection methods (see \textit{Articles}) \cite{castelo_agnostic_2019}. The common assumption, across disciplines, is that coordinated networks leave more detectable evidence of manipulation than do individual accounts, and that these footprints should be identifiable regardless of attack type or rumoring topic \cite{castelo_agnostic_2019,milajerdi2019holmes}. Network-scoped methods, including relevant security literature, form 20\% of our corpus.

\circled{1} \textbf{Detection targets.} \textit{All} methods at this scope identify patterns of user interaction and content propagation as targets; these methods associate virality with the existence of rumoring narratives \cite{vosoughi-twitter-2018,tambuscio_fact-checking_2015,shu_beyond_2019} and temporally anomalous activity with evidence of coordination \cite{stone2009botnet,cresci_bots_2020,lukito2020coordinating}. The corresponding targets for these approaches are \hl{\textbf{(N.i)}} \textit{propagation patterns across social graph topologies} and \hl{\textbf{(N.ii)}} \textit{temporal records of user-user interactions}. Within non-security misinformation literature, we note that virality assumptions disallow \textit{early} detection of misinformation \cite{monti2019fake,castillo_information_2011}. Similarly, within the security literature, anomalous patterns of account registration and user interaction serve as proxies for the presence of Sybils and botnets \cite{yuan2019detecting,danezis2009sybilinfer}; early detection requires that authors formulate a priori assumptions about the nature of these patterns.

\circled{2} \textbf{Datasets.} 
 We conducted an author outreach survey for works within this scope in an attempt to locate hard-to-find social media datasets. We found that \hlcyan{\textbf{(2.v)}} \textit{accessibility issues} were exacerbated by the shutdown of the Twitter API \cite{twitter}. In total, we attempted to locate datasets and code for 50 different papers (see~\Cref{sec:rep_studies} for our methodology). Thirty-six (72\%) of these analyzed tweet
corpora, and 42 (84\%) of these targeted social media users and posting contents. We were able to independently source complete methods or data for fourteen (28\%) of these. Of the 27 authors we eventually contacted about providing partial or dehydrated datasets, nine responded; six of those authors were able to provide method code or partial datasets.

\circled{3} \textbf{Model selection.} The network-scoped methods we review formulate detection as 1) a structured content classification problem \cite{alizadeh2020content, ratkiewicz2011detecting}, and/or 2) a clustering problem on social graphs \cite{assenmacher2020two, sharma_combating_2019}. In the former case, authors employ an assortment of \hlred{(\textbf{3.ii, 3.iii})} \textit{supervised and unsupervised models} to detect suspicious language across multiple accounts. In the latter case, authors use \hlred{(\textbf{3.vi})} \textit{Louvain or K-means clustering or K-nearest neighbors} to detect neighborhoods of suspicious accounts, as determined by user-user interactions. Though methods in the latter category advertise themselves as content-agnostic, we note that published methods \cite{liu_early_nodate, ma_detect_2017} access datasets of social media posts that were already sorted by rumoring topic or event \cite{ma_rnns_2016, liu_realtime_2015}.

\circled{4} \textbf{Feature selection.}
55\% of papers within this scope make normative theoretical assumptions about user behaviors: In keeping with an epidemiological model\footnote{Some academics have argued that the disease metaphor for misinformation promotes an overly simplistic model of information spread and opinion formation \cite{williams_2023, tay2024thinking}.}  of misinformation spread \cite{jin_epidemiological_2013, vanderlinden_2022}, Nguyen et al. assume a homogeneous population of newsreaders, with \hlgreen{\textbf{(4.ii)}} \textit{identical probabilities of ``infection'' and reinfection} \cite{nguyen_containment_2012}. Similarly, in the security literature, techniques for detecting bots and Sybils identify behaviors that align with heuristics determined a priori by researchers: These methods assume, for instance, that Sybils will form well-connected neighborhoods \cite{yuan2019detecting}, or (seemingly contradictorily) that compromised Sybils will refrain from connecting with additional Sybils, to avoid detection \cite{danezis2009sybilinfer}.\footnote{In fact, the landmark paper by Douceur that characterized Sybil attacks stated that such attacks cannot be prevented unless special assumptions are made about account behaviors \cite{douceur2002sybil}.} 

\circled{5} \textbf{Evaluation.} Ferrara et al. \cite{ferrara2016rise} called attention to the false positive rate problem in botnet detection in 2016, noting that classifiers for bot detection only work well in instances where there is a clear-cut distinction between bot and non-bot accounts. This distinction is becoming increasingly blurred, however. Sophisticated network-based attacks try to engage non-bot accounts in organic interactions with bot accounts (e.g., astroturfing attacks) \cite{cresci_bots_2020,stone2009botnet}, \hlorange{\textbf{(5.v)}} \textit{rendering even positive detection results insufficient or meaningless}: coordinated activity need not be inauthentic, and inauthentic activity need not be malicious.

% \begin{tcolorbox}

\textbf{**Takeaways}: \circled{1} Pattern- and virality-based detection approaches disallow early detection of rumors. \circled{2} Current social media data is difficult to obtain. \circled{3} ``Topic-agnostic'' classifier design occurs downstream of topic-aware dataset design. \circled{4} Epidemiological models of information spread make strong assumptions about opinion formation and user behaviors; feature sets reflect these \textit{a priori} notions. \circled{5} Though network-scoped methods are less susceptible to content-based dependencies than are content-aware methods, they cannot infer intent or authenticity of the behaviors they detect.
% \end{tcolorbox}

\subsection{Websites} 
\label{sec:websites}
Methods within this scope apply credibility, factuality, or (political) bias scores to whole news sites; authors claim that site-wide labels can be used to quickly infer the quality of individual news articles produced by these sites \cite{hounsel-infra-2020,baly-etal-2020-written,castelo_agnostic_2019,chen_proactive_2020}. As with article-scoped methods, we noted intervention fit issues at the whole website scope. Asr et al. found that whole-source labels were insufficient proxies for the truthfulness of single news articles, and elided subject-specific variations in reporting quality \cite{torabi_asr_data_2018}. (We discuss dataset distribution in greater depth in \textit{Datasets}.) 

\circled{1} \textbf{Detection targets.}
In 50\% of works that we review in this scope, authors reduce the task of whole-site credibility labeling to a significantly smaller, unimodal classification task: Chen et al. \cite{chen_proactive_2020} \hl{\textbf{(W.i)}} \textit{detect suspicious domain semantics} (in essence, a text classification task on URLs); Ribeiro et al. \cite{ribeiro_media_2018}  \hl{\textbf{(W.ii)}} \textit{infer site bias from site visitor demographics}; Castillo et al. \cite{castillo_information_2011} \hl{\textbf{(W.iii)}} \textit{detect suspicious ad interfaces and markup features}; Baly et al. \cite{baly_predicting_nodate, baly-etal-2020-written} and Hounsel et al. \cite{hounsel-infra-2020} present methods incorporating \hl{\textbf{(W.iv)}} \textit{infrastructural features}, though the overall performance of the method of Baly et al. is strongly determined by performance on the text classification task alone (thus, in practice, the method closely resembles the article-scoped detection methods we examine, and is susceptible to the same dependencies that we observe in that scope). We demonstrate this via an ablation analysis in our replication study of their method (see~\Cref{sec:baly_rep}). 

\circled{2} \textbf{Datasets.} For both training and testing, all methods scoped to whole website detection rely on published lists of websites with accompanying credibility scores \cite{baly_predicting_nodate, baly-etal-2020-written,hounsel-infra-2020,castelo_agnostic_2019}; common reference sites include Media Bias/Fact Check, Snopes, and FactCheck.org \cite{mediabiasfactheck,snopes,FactCheck}. As discussed in~\Cref{sec:claims}, however, these references do not have uniform taxonomies for classifying site credibility. Additionally, pre-labeled lists are 1) \hlcyan{\textbf{(2.i)}} \textit{biased towards older, more visible real news and fake news outlets} \cite{snopes}, 2) \hlcyan{\textbf{(2.iii)}} \textit{are restricted to specific information domains} \cite{baly_predicting_nodate}, or 3) \hlcyan{\textbf{(2.i)}} \textit{include inactive websites within their labeled datasets} \cite{mediabiasfactheck}.\footnote{The median lifespan of a set of 283 misinformation news sites is 4 years, per a survey conducted by Chalkiadakis et al. in 2021 \cite{chalkiadakis_rise_2021}.} We note papers that perform website infrastructure analysis on contemporaneous snapshots (circa 2019) of websites in their corpora, even though the text-based features for the same analysis were drawn from datasets published in 2016 and 2017 \cite{castelo_agnostic_2019}. This constitutes a serious source of \hlcyan{\textbf{(2.ii)}} \textit{temporal leakage}. No works within this scope discuss approaches to accounting for uneven distributions in training data, or how they might account for shifts in baseline distributions during the lifecycle of an active website. Hounsel et al., for instance, train their classifier on a reference list in which 34\% of misinformation training set sites were active and \textit{all} websites in the real news training set were active, possibly resulting in overfitting to features specific to those inactive websites \cite{hounsel-infra-2020}. 

\circled{3} \textbf{Model selection.} Four of the seven methods we reviewed within this scope employed \hlred{(\textbf{3.ii})} \textit{SVM classifiers}; in two of those cases, SVM outperformed other, more complex unsupervised models \cite{torabi_asr_data_2018, castelo_agnostic_2019}. These results accord with our earlier observation, in~\Cref{sec:articles}, that SVM classifiers are comparatively performant on two-way classification tasks. 

\circled{4} \textbf{Feature selection.} Works within this scope employ multi-modal feature sets comprising a mix of \hlgreen{\textbf{(4.i)}} \textit{textual}, \hlgreen{\textbf{(4.ii)}} \textit{network-based}, and \hlgreen{\textbf{(4.iv)}} \textit{infrastructural signals}: for instance, Baly et al. \cite{baly-etal-2020-written} consider network traffic, URL semantics, and site contents; and Hounsel et al. \cite{hounsel-infra-2020} consider TLS/SSL certificates, web hosting configurations, and domain registrations. None of the detection methods we reviewed discusses the computational costs of deploying their methods at scale. Though all works discuss their feature selection process (via leave-one-out and use-one-only evaluations), none describes a process for normalizing or weighting features according to dataset distribution or detection setting needs. 

\circled{5} \textbf{Evaluation.} Methods within this scope that propose to perform whole-site labeling from analysis of a selection of news articles or infrastructural features are susceptible to distributional imbalances (e.g., between news verticals represented in an article corpus). Baly et al. train their model on \hlorange{\textbf{(5.ii)}} \textit{political news websites only}, and their credibility labels are strongly correlated with political bias scores. Hounsel et al. perform \hlorange{\textbf{(5.i)}} \textit{testing in real time}---one of the few studies we reviewed, and one of two studies at this scope, that did so \cite{hounsel-infra-2020, chen_proactive_2020}. Both of these works report significant performance dropoffs between experimental and real-time tests, most likely as a result of distributional differences between real-world and experimental datasets (in practice, most websites do not host news-related content at all) \cite{hounsel-infra-2020, chen_proactive_2020}.  

\textbf{**Takeaways}: \circled{1} Methods claiming to classify news sites generally reduce this task to simpler, unimodal ones, such as URL classification. \circled{2} The works we review do not consider shifts in feature distribution over time. \circled{3} We note that SVM classifiers perform well on two-way classification tasks, and even outperform more sophisticated unsupervised models. \circled{4} Feature normalization largely undiscussed. \circled{5} Authors who conduct testing in separate real-time settings report significant performance dropoffs with respect to experimental results.

\section{REPLICATION STUDIES}
\label{sec:rep_studies}

We choose three distinct targets and scopes to replicate issues identified in our literature review. Our targets are 1) \hl{(\textbf{A.i})} syntax-based text features; 2) \hl{(\textbf{U.ii})} user behaviors; and 3) \hl{(\textbf{W.iv})} multimodal whole-website features, including hosted content and infrastructure. These works are highly representative of their respective information scopes: Nasir et al. \cite{nasir_fake_2021} (\Cref{sec:article_rep}) train a neural network on corpora of true and misinformative news articles; Saeed et al. \cite{saeed2022trollmagnifier} (\Cref{sec:trollmagnifier}) use published lists of trolls to infer the presence of other troll accounts on Reddit; and Baly et al. \cite{baly_predicting_nodate} (\Cref{sec:baly_rep}) examine a multimodal feature set comprising infrastructure, content, and network-based features in order to infer whole-site credibility. For each work, we evaluate replicability and generalizability (\textcolor{blue}{RQ3}, \textcolor{blue}{RQ4}). Where possible, we 1) replicate reported results; 2) inspect datasets for potential dependencies; 3) perform ablation analyses to understand individual feature performance; and 4) test on current data. 

\textbf{Paper selection criteria and author outreach.} We sorted our full text corpus by information scope. Within each scope, we sorted papers in order of decreasing citation count. We then proceeded as follows:

\begin{enumerate}[nolistsep]
    \item[1.] We attempted to source the full methods and datasets for the most cited paper within each information scope.

    \item[2.] If we were unable to find this information during an independent web search, we reached out to the paper's lead author(s) to request access.

    \item[3.] If this request was unsuccessful---if the author did not respond, or confirmed that the dataset or code was no longer available---we returned to step 1 for the next most highly-cited paper in our corpus within that scope.
\end{enumerate}  

\textbf{Replication analyses.} In order to reproduce results published in a selection of papers and perform cross-cutting analyses on out-of-domain and out-of-sample datasets, we conduct a series of replication analyses on a subset of papers. We chose representative methods from disparate information scopes, and which consider a variety of different feature types. We reproduced results reported in each paper on the datasets mentioned therein, contacting authors when necessary to obtain datasets and code. We evaluated reproducibility and generalizability as follows:

\begin{itemize}[nolistsep]

    \item \textbf{Reproducibility.} We reproduce published results with code and data reported in the original publication. We evaluate availability of code and data and, where possible, compare our analysis outcomes with those reported in the original paper (\textcolor{blue}{RQ3}). 

    \item \textbf{Explainability.} Toward understanding the contributions of specific feature types to overall classifier performance, and why certain approaches work, we perform feature ablation studies when appropriate. 

    \item \textbf{Replicability and generalizability.} Model perfor-mance on novel datasets is useful for determining the generalizability of existing detection methods to different contexts (\textcolor{blue}{RQ4}). For models that were explicitly tested on specific misinformation narratives (e.g., 2020 stolen election narratives), on specific timeframes, or on specific types of misinformation (e.g., parody, satire), we develop updated datasets to test method performance on diverse information domains. 
\end{itemize}

%--------------------------------------------------------------

\subsection{Detection of suspicious language}
\label{sec:article_rep}
In view of rampant data dependency issues identified in our survey of article-scoped literature, we reproduced results from a representative study published in 2021 by Nasir et al \cite{nasir_fake_2021}. The authors propose a neural net-based approach to the classification of \textit{news articles}. The method employs a hybrid deep learning model that combines convolutional and recurrent neural networks for the classification of real and fake news. The authors report results for tests on two datasets: the ISOT dataset (45,000 news stories, equally distributed across true and false categories, as labeled by PolitiFact) and the FA-KES dataset (804 news stories about the Syrian war, 426 true and 376 false) \cite{isot, fa_kes}. We were able to replicate original paper results and run the method on updated datasets of news articles. Motivated by the prevalence of methods trained on few- or single-source datasets in the article scope, we test possible dependencies related to journalistic house style, as \textit{all true articles in the training corpus were sourced from Reuters.}

\textbf{House style as a confounder.} To investigate house style as a possible confounder for misinformation detection, we excerpted 100 articles from both Reuters and The New York Times, two news outlets with distinctive (and different) reporting styles. We randomly selected these articles from both outlets' RSS feeds in May 2023. We sourced articles for both corpora from the following verticals: U.S. and world politics, economics, science, and entertainment. Excerpt lengths ranged from 100 to 300 words. These corpora, each comprising 100 true news stories, are labeled ``Reuters-\texttt{original}'' and ``NYTimes-\texttt{original}'' in Table~\ref{tab:nasir}.

\begin{table}[!t]
\small
\centering
\def\arraystretch{1.3}
\begin{tabular}{llrrr}%
\toprule
\textbf{Dataset (size)} & \textbf{Features} & \textbf{Acc.} & \textbf{FPR} & \textbf{FNR}\\
\midrule
\multirow{2}{2cm}{ISOT (45,000)} & 
        \texttt{original} & 0.995 & 0.00 & 0.00 \\    
    &   \texttt{scrubbed} & 0.987 & 0.01 & 0.00 \\ 
\midrule
FA-KES (804) & \texttt{original} & 0.521 & 0.151 & 0.843 \\ 
\midrule
\multirow{2}{2cm}{Reuters (100)} & 
        \texttt{original} & 0.727 & 0.71 & -- \\ 
     &  \texttt{modified} & 0.507 & 0.02 & 1 \\
\midrule
\multirow{2}{2cm}{NYTimes (100)} &
         \texttt{original} &  0.673 & 0.705 & -- \\ 
       & \texttt{modified} &  0.492 & 0.735 & 0.286 \\ 
\midrule
ChatGPT (250) & 
         \texttt{original} & 0.939 & 0.02 & -- \\
\bottomrule
\end{tabular}
\caption{Replication analysis of Nasir et al. (2021). We tested the method of Nasir et al. on both datasets discussed in the original paper, and on novel datasets from Reuters and The New York Times \cite{nasir_fake_2021}.} 
\label{tab:nasir}
\end{table}

We then selectively edited 50\% of the news articles within each corpus. We changed proper nouns, negated verbs, and altered reported statistics so that the factual content of these articles was no longer accurate but house style and tone were preserved. We call these altered text corpora ``Reuters-\texttt{modified}'' and ``NYTimes-\texttt{modified}'' in Table~\ref{tab:nasir}. The classifier had 0.727 accuracy on Reuters-\texttt{original} and 0.673 accuracy on NYTimes-\texttt{original}; this difference is not significant (\textit{p} > 0.05). Additionally, the classifier had about 0.50 (random) accuracy on both modified datasets. The classifier's false positive and false negative rates on modified and original Reuters and NYTimes corpora tell a more interesting story, however: Overwhelmingly, \texttt{false} NYTimes articles were classified as \texttt{true} (FPR = 0.02 and 0.735 for Reuters-\texttt{modified} and NYTimes-\texttt{modified}, respectively), while all \texttt{true} Reuters articles were classified as \texttt{false} (FNR = 1 and 0.286 for Reuters-\texttt{modified} and NYTimes-\texttt{modified}, respectively). These results indicate that the classifier was significantly more conservative in its assignment of \texttt{true} labels for the Reuters dataset than it was for the NYTimes dataset; additionally, \textit{within} each modified corpus, the classifier did not effectively differentiate between true and untrue news articles. We note that classifier performance in an adversarial setting was no better than random, and that house style appears to have a significant impact on classifier sensitivity to misinformative texts.

%------------------trolls/networks---------------------

\subsection{Detection of suspicious accounts}
\label{sec:trollmagnifier}
 We reproduce results from a study published in 2022 by Saeed et al. \cite{saeed2022trollmagnifier}. In summary, the authors propose a method, called TrollMagnifier, for the identification of Reddit accounts that exhibit troll-like behaviors. Like all other account-scoped methods we analyze in the security literature, TrollMagnifier is trained on posting and reply statistics for non-troll and known Russian troll accounts (as identified by Reddit) \cite{saeed2022trollmagnifier}.

\textbf{Reproducibility of published results.} Study authors provided us with pre-processed datasets and classifier code upon email request. The full Reddit Pushshift dataset is freely available online \cite{baumgartner2020pushshift}. We were able to replicate original paper results using these materials. Original account handles were anonymized; as such, we were unable to verify if the accounts identified in the original study appeared to be troll-like. 

\begin{figure}[!t]
\centering
% \makebox[\textwidth]
\makebox[0pt]{\includegraphics[scale=0.53]{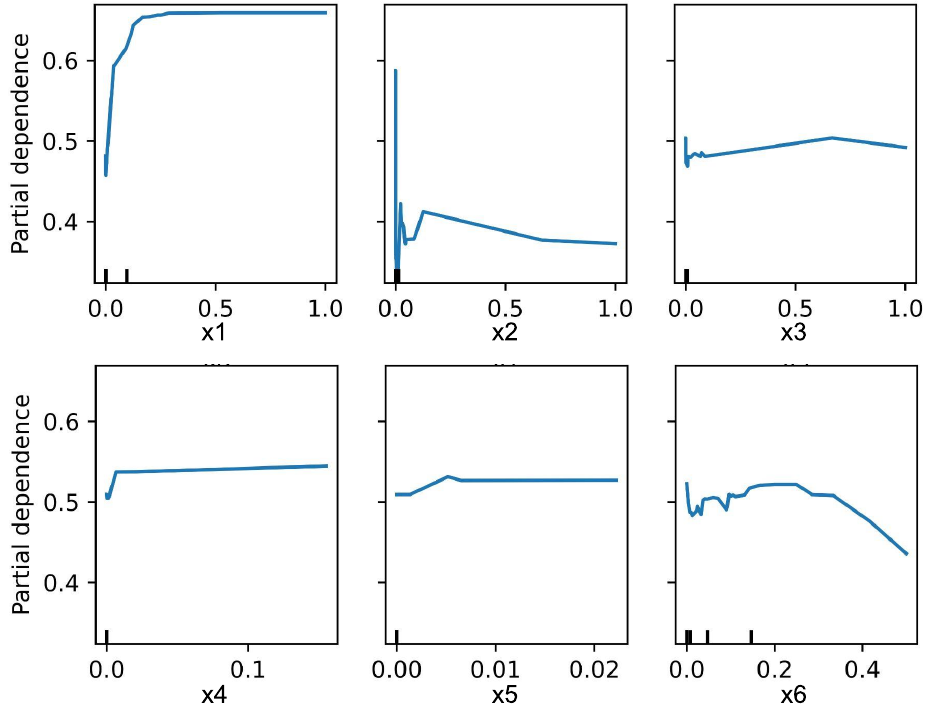}}
\label{fig:saeed_pdp}
\caption{Partial dependence plots for each TrollMagnifier feature. Respectively, x1 = ``comments on posts that trolls commented on,'' x2 = ``comments on posts that trolls started,'' x3 = ``direct comment in reply to troll post,'' x4 = ``threaded comment in reply to troll comments on a troll post,'' x5 = ``same title post as troll,'' x6 = ``same title post as troll.''}
\end{figure}

\textbf{Tautological targets.} 
As described in preceding sections (\Cref{sec:accounts}), suspicious account detection suffers from a lack of clear and consistent definitions. Troll accounts cannot be described by degree of automation (while some trolls are bot-like, many others are operated by humans) \cite{cresci_bots_2020} or the nature of the information they spread (this might be political misinformation, ads, or anything in between) \cite{yang2022botometer}; as such, the clearest definition that Saeed et al. implicitly offer is that a troll is an account that exhibits \textit{troll-like behavior}: i.e., interacts with known troll accounts, or appears on the same posts or message threads as these accounts. The authors note in their own work that this approach cannot be used to detect novel trolling behaviors, and requires a seed list of known troll accounts for every new detection task. 
% (we note in our discussion of \textit{Account}-scoped data curation practices that the process of identifying these seed accounts is actually a fairly manual one). 

\textbf{Implied versus actual data dimensionality.} While the original TrollMagnifier paper strongly implied that the proposed method would leverage networked behaviors to identify troll accounts acting in coordination online, we found that, in actuality, the features under analysis lacked any sort of temporal component and were limited in scope. There were six features in all (described in full in Figure~2), each corresponding to an aggregate engagement statistic. Longitudinal data and timestamps were not available; as such, it was not possible to perform time series analysis. Account names were not available, disallowing construction of user graphs.

\textbf{Feature importance.} In our partial dependence analysis, we find that feature x1---commenting statistics---was the sole feature that consistently produced classification accuracy greater than 0.6 (most other features had accuracy no better than random). Per our earlier observation that many account-scoped methods target behaviors that are difficult to distinguish from routine online activity, we recommend that feature engineering for account- and network-scoped methods reflect some intuition about the nature of actually suspicious behaviors. Furthermore, the performance of the current feature set suggests that manual classification might be as effective as (or even more effective than) an automated approach that detects a content-agnostic heuristic. 
%------------------websites---------------------

\subsection{Detection of suspicious websites}
\label{sec:baly_rep}
 We reproduce results from a study published in 2018 by Baly et al \cite{baly_predicting_nodate}. In summary: the authors propose a multimodal approach to the classification of \textit{news websites}. This method is particularly representative of works within this information scope: 50\% of works within this scope use a similar mixed-modalities approach to detecting misinformation websites, and Baly et al. include site-specific feature types, including domain and traffic-based features, in their analysis. Baly et al. analyzed website contents, associated social media accounts, and Wikipedia pages in order to perform two classification tasks: fact and political bias classification. The authors developed a dataset of 1066 websites manually labeled for their political leaning (extreme-left, left, left-center, center, right-center, right, and extreme-right) and degree of credibility (low, mixed, high). These labels were extracted from the Media Bias/Fact Check (MBFC) database \cite{mediabiasfactheck}. We were able to reproduce original study results and perform ablation analyses on existing datasets. We were unable to run the method on an updated dataset, as feature extraction code was not available. 

\textbf{Reproducibility of published results.}
All features extracted for the original analysis were captured in a series of json files. While we were able to readily reproduce results reported in the paper, certain elements of the dataset (follower counts on social media, Wikipedia page contents) were out of date. As we did not find documentation in the method repository for re-extraction of these features, we were restricted to conducting our tests on data that were already available. We binned bias labels into \textit{left}, \textit{center}, and \textit{right} categories, as the seven-way taxonomy initially applied to the dataset by MBFC yielded small label classes.  Classifier performance on the resulting three-way bias classification task accords with the results reported by Baly et al. on the same task.

\textbf{Multimodal features: help or hindrance?} 
We performed an ablation study of the method on the EMNLP18 dataset and analyzed the method's \textit{bias} and \textit{fact-checking} classification functions separately \cite{baly_predicting_nodate}. Specifically, we stratified the original EMNLP18 dataset by political leaning and credibility, as labeled by MBFC, and analyzed the performance of 1) the full feature set, 2) individual features and 3) ablated feature sets (removing one feature type per test). Our results are summarized in~\Cref{tab:baly_ablation}. We find that, on 11 out of 12 test datasets, classifier performance using only text-based features (\texttt{articles} and \texttt{wikipedia}, derived from articles randomly sampled from the website in question, and the site's corresponding Wikipedia page, respectively) was comparable to performance on the full feature set. On five out six datasets, bias classification accuracy on text-only features actually outperformed bias classification on the whole feature set (see the bottom half of~\Cref{tab:baly_ablation}), suggesting that the full-site classifier of Baly et al. was effectively a text content classifier.

\section{DISCUSSION}
\label{sec:discussion}
 We focus our discussion of results on those issues identified by our literature review and investigated in greater depth in our replication analyses. Additionally, we provide recommendations for evaluating ML-driven trust and safety interventions and link these recommendations to major takeaways of the present study.

\textcolor{blue}{RQ1}: \textbf{Fit.} Very few methods that claimed to detect misinformation performed actual fact verification: Instead, they targeted proxy signals that were frequently steps away from promised detection targets. These differences were particularly noticeable in methods that relied heavily on text- and network-based features to perform classification. In those cases, semantic/syntactic signatures and propagation patterns served as proxies for the existence of misinformation. We demonstrated, through our own replication studies, that it is easy to circumvent approaches that rely on style-based cues to perform proxy detection. 

\textcolor{blue}{RQ2}: \textbf{Data curation and model explainability.} Lack of access to current, well-annotated datasets remains a serious problem for current and future misinformation research. Across existing datasets, taxonomies for classifying misinformation were inconsistent. Testing on contemporaneous data was uncommon among those papers we annotated, and testing in real-time settings was even rarer. Proof-of-concept experiments oftentimes did not control for data dependencies. Authors describing black-box methods---particularly those employing neural nets or other forms of unsupervised learning---did not disclose feature sets retrieved by their methods. 

\textcolor{blue}{RQ3}: \textbf{Reproducibility.} We noted widespread code and data availability issues: Fewer than 30\% of our attempts to locate code \textit{and} data, or obtain this information from authors, were successful. The code and datasets we \textit{were} able to retrieve were frequently unusable or out of date. In fact, we were able to reproduce and replicate results on published \textit{and} current data for only one of our replication studies. In that single case, we found that the article-scoped method performed no better than random (0.50 accuracy) on a current, mixed-domain dataset (\Cref{sec:article_rep}).

\textcolor{blue}{RQ4}: \textbf{Generalizability.} Methods that were trained on single-source or single-domain datasets appeared to perform well on data from the same source, or within the same domain; these methods, unsurprisingly, performed poorly on multi-source or out-of-domain topics. These discrepancies are closely tied to undisclosed or uncontrolled-for data dependencies, which we discuss in RQ2.

\subsection{Recommendations for future research}
\label{sec:rubric}
We propose the following as directions for future research: 

\textbf{Designing for intervention fit.} \textit{All article-scoped methods detect proxy targets for misinformation. Some of these proxy targets, including house style and text sentiment, are easily circumvented (\Cref{sec:article_rep}).} At the point of task formulation, researchers might consider the following: What signals---or proxy signals---will the intervention detect? What are the potential risks of a false negative or false positive result (and are they worth the potential benefits of e.g., greater speed)? What is the actual detection target, and how closely does it approximate the classification task to be performed? In an adversarial setting, would a malicious actor be able to circumvent the automation?

\textbf{Understanding hybrid detection.} \textit{While academic misinformation detection favors fully automated methods, commercial checking services and online platforms employ hybrid methods.} Increasingly, online services are turning toward human-driven hybrid detection approaches, wherein human users make complex fact-checking decisions and automated methods amplify their decision-making power (e.g., by identifying content similar to already-flagged posts). We believe that exploration of the interplay between human and automated decision-making systems will prove a fruitful frontier for academic research. And, in fact, recent work \cite{sehat2024misinformation, lykouris2024learning} frames the hybrid detection problem as one of triage and scheduling: how can automated approaches quickly surface urgent cases for further review by human moderators?

\textbf{Investigating distributional shifts.} \textit{Fewer than 10\% of methods within our whole corpus tested on contemporaneous data, and fewer than 5\% tested in a real-time setting.} To understand how an ML intervention might work \textit{in practice}, researchers must understand how robust their methods are to changes in the distribution of available data during training versus testing. These shifts are particular to medium (for instance, whole site classifiers are susceptible to fluctuations in news vertical coverage and author). Notably, the few works in our corpus that tested in real-time reported precipitous performance dropoffs.

\textbf{Addressing nuanced detection challenges.} \textit{We note that, in general, the detection methods we consider in this work elide subtleties that are particular to the medium under consideration}: for instance, article-level detection methods generally apply broad `true'/`false' classifications to a text under analysis where only single sentences might be slightly inaccurate. Additionally, the language signals that most methods expressly detect are fairly unsubtle, and are likely verifiable via manual means. Suggestion, insinuation, and leading questions are powerful rhetorical tools that might render a newsreader more susceptible to actual misinformation, or that might suggest misinformative ideas via indirect means; no works within our literature corpus expressly targeted these forms of language, however.

%-----------------------------------------------------
\section{Ethics considerations}
\label{sec:ethics}

Our author outreach survey design was approved by the Princeton University IRB. All statistics published from those communications are in aggregate, with no personal identifiers attached to individual author responses. Datasets used for this work were already publicly available or were obtained with permission from study authors.

\section{Open science}
\label{sec:open_science}
Code and data used for our analyses, as well as our full paper corpus, are included in our online SI (available at \url{https://anonymous.4open.science/r/sok_misinformation-41E8}). 

\section*{Acknowledgments}
The authors would like to thank Anne Kohlbrenner for assisting with paper coding; Ben Kaiser and Sarah Scheffler for productive conversations, and for reviewing multiple drafts of this manuscript; and Mona Wang for retrieving data used for a replication study. \\

\nocite{*}

\printbibliography

\appendix
\label{appendix}
\section*{Appendix}

\section{Commercial fact-checking services}
\label{sec:commercial_llm_checking}
We include here a brief market survey of commercial and LLM-powered fact-checking and IO detection services. In general, these services fall into five categories: 1) media fact-checking organizations; 2) brand safety and suitability services; 3) trust \& safety operations at large social media platforms, 4) threat detection operations, and 5) analytics organizations unaffiliated with a media outlet that offer research capacity to governments and businesses. We define each service category and (with the exception of the first category, which comprises human media workers and fact-checkers) discuss automated content moderation operations deployed by three prominent exemplars within each service category. In general, in instances where such information is made available, we observe that at-scale content moderation businesses \textit{at least} employ human-labeled datasets to train classifiers, and some retain subject-area experts to adjudicate complex moderation decisions. On social media platforms, in particular, human moderators and automated systems appear to work hand-in-hand: automated systems surface potentially misinformative content that receives final verification from a human moderator. For IO detection, specialized knowledge (pertaining to specific geographies, languages, or political climates) is often invoked.

\textbf{Media fact-checking.} Human fact-checkers and content moderators affiliated with news outlets, or who work as freelance fact-checkers. \textit{The International Fact-Checking Network (IFCN)} is a professional network of media workers and fact-checkers; IFCN is also the de facto standards setting body for media fact-checking, and maintains a fact-checking code of ethics \cite{ifcn_code}. In general, media fact-checking organizations with IFCN affiliations are established news organizations, non-profits, and watchdog organizations that employ human journalists and fact-checkers. Furthermore, (human) fact-checkers can receive IFCN compliance certificates after passing a qualifying exam. 

\textbf{Brand safety and suitability companies.} B2B companies that detect categories of potentially harmful speech on websites where ads might appear. Advertisers wishing to protect ``brand safety'' contract with these services to ensure that their ads do not appear alongside problematic content. The Global Alliance for Responsible Media (GARM) is the standards-setting body for brand safety and suitability companies \cite{garm_standards}. 
\begin{itemize}
    \item \textit{Zefr}, a GARM member company, deploys AI to detect material that falls within predefined subcategories of problematic content (e.g., explicit content, misinformation, spam). In a press release for Zefr's acquisition of an AI-driven content moderation company (AdVerif.ai) from 2022, the company disclosed that AdVerif.ai is ``powered by fact-checking data from more than 50 IFCN-certified organizations around the globe'' \cite{Zefr_2022}---that is, AdVerif.ai trains its models on labeled datasets produced by (human) IFCN affiliates.

    \item \textit{DoubleVerify}, a GARM member company, ``uses sophisticated approaches that rely on a combination of AI and comprehensive human review'' \cite{Pasquine_2022}. According to the company's documentation, human assessors (a ``semantic science team'') evaluate site infrastructure and contents; AI is used to scale their assessments. 

    \item \textit{Integral Ad Science} (IAS), a GARM member company, deploys AI to detect low-quality sites via infrastructure features. The company's data sources, and deployment methodology were not immediately evident upon web search; IAS recently announced a new partnership with Meta for ad placement management on Facebook \cite{Integral_Ad_Science_2024}.
    
\end{itemize}

\textbf{Trust \& safety operations.} In-house content moderation teams at large social media platforms. 
\begin{itemize}

    \item \textit{Twitter} has deployed a crowd-sourced annotations platform called Community Notes (formerly Birdwatch) since 2021 \cite{wojcik2022birdwatch}. 

    \item Until recently, \textit{Facebook} partnered with IFCN affiliates to perform third-party manual checking of possibly misinformative content; first-line automated methods detect potentially harmful speech and surface near-duplicates of known problematic image (SimSearchNet++) and text content  \cite{facebook_files_2021, fb_third_party, AI_at_Meta_2020}. Meta announced in January 2025 that it was sunsetting its third-party checking program in favor of a Community Notes-like system \cite{chan_meta_2025}.
        
    \item  \textit{TikTok} employs thousands of content moderators across the globe who ``work alongside automated moderation systems'' \cite{McIntyre_Bradbury_Perrigo_2023, Farah_2023}.

\end{itemize}

\textbf{Threat intelligence services.} At-scale detection of advanced persistent threats, foreign influence operations, and other cyberattacks oftentimes perpetrated by nation state actors. 

\begin{itemize}

    \item \textit{Mandiant} strongly implies the use of hybrid detection methods, and disclaims that ``defenders must constantly explore different techniques and leverage both subject matter expertise and technical capabilities to filter and uncover malicious activity'') \cite{Kapellmann_Zafra_Serabian_Riddell_Brubaker_2022}.   

    \item \textit{Microsoft Threat Intelligence} strongly implies the use of hybrid detection methods; in a report from September 2023, MTI cites the work of in-house ``Microsoft Security teams'' which are tracking an advanced social engineering attack \cite{Microsoft_Incident_Response_2023}. Other details---including possible use of automated methods---are undisclosed.

    \item \textit{Facebook Coordinated Inauthentic Behavior} reports share quarterly updates about Meta's takedown of coordinated activities across its platforms \textit{and} others, including local news outlets. In a report from February 2023, Meta describes a CIB network in Serbia that used local news media to create the impression of grassroots support for the Serbian Progressive Party; while the nature of the detection  methodology is unspecified, the complexity and geographic specificity of the CIB described suggest that specialists with country-level expertise were likely consulted \cite{Nimmo_2023a}. 
\end{itemize}

\textbf{Analytics firms.} For- and non-profit organizations that offer checking services and research capacity to governments and businesses. 

\begin{itemize}
    \item \textit{The Global Disinformation Index (GDI)} ``reviews news domains based on various metadata and computational signals.'' Content, however, is manually reviewed by a ``country expert,'' who analyzes a random sample of 10 articles from a news site to determine veracity \cite{Srinivasan_GDI}. 

    \item \textit{DFRLabs (Digital Forensic Research Lab)} has disclosed that it employs human subject-area experts, and primarily addresses technology and policy issues pertaining to global and international affairs. In 2018, Facebook contracted its services to detect online trolls \cite{Lapowsky_2018}. 

    \item \textit{Graphika Labs} leverages network analysis to identify influence operations online. On its own website and in the popular press, Graphika has disclosed that it uses AI to map online networks and trace information flows \cite{Schiffrin_2023, graphika_ai}.  
    
\end{itemize}

\textbf{LLM-driven detection.} A few LLM-powered detection methods have been discussed in the popular press, including those advertised by Google \cite{Storey_2023} and OpenAI \cite{fried_2023}, but these deployments appear to be mostly experimental, or have required additional adjudication from human moderators. OpenAI in particular has advertised content moderation tools that address misinformation-adjacent tasks, such as toxic speech detection \cite{perspective_openai}. Misinformation and toxic speech detection are not equivalent tasks, however, and the latter is narrowly defined in the Perspective training data documentation as a four-way classification task (the four class labels are ``profanity/obscenity,'' ``identity-based negativity,'' ``insults,'' and ``threatening'' language). 

% \section{Paper-by-feature summary}
\begin{table*}[htbp!]
\centering
\resizebox{!}{4.1in}{
    \begin{tabular}{lc|c|c|c|cccc|r}
    \multicolumn{2}{c|}{\textbf{\ul{Paper}}} &
    \multicolumn{1}{c|}{\circled{1} \textbf{\ul{Target}}} &
    \multicolumn{1}{c|}{\circled{2} \textbf{\ul{Dataset}}} & \multicolumn{1}{c|}{\circled{3} \textbf{\ul{Model}}} & \multicolumn{4}{c|}{\circled{4} \textbf{\ul{Features}}} & \multicolumn{1}{c}{\circled{5} \textbf{\ul{Performance}}} \\
    \textbf{Work} & \textbf{Scope} &  &  &  & \begin{sideways}{\textbf{Textual}}\end{sideways} & \begin{sideways}{\textbf{Network}}\end{sideways} & \begin{sideways}{\textbf{Author}}\end{sideways} & \begin{sideways}{\textbf{Infra.}}\end{sideways} &  {\textbf{Accuracy/AUROC}} \\
    \hline\hline
1. Ajao et al. \cite{ajao_sentiment_2019} & \circled{C} \circled{A} & Sentiment \textbf{(A.i)} & PHEME \cite{pheme_2018} & LSTM, DT, RF, \underline{SVM} & \fullcirc &  & & & 0.86 (Acc.) \\

2. Abulldah-All-Tanvir et al. \cite{abdullah-all-tanvir_detecting_2019} & \circled{C} \circled{A} \circled{N} & Content \textbf{(C.i)} & Twitter (API) & NB, RNN, LSTM, \underline{SVM}, Logit & \fullcirc & \fullcirc & & & 0.89 (Acc.) \\

3. Bhutani et al. \cite{bhutani_fake_2019} & \circled{C} \circled{A} & Content \textbf{(C.i)}; sentiment \textbf{(A.i)} & Twitter (API), PolitiFact \cite{politifact} & Naive Bayes, \underline{RF} & \fullcirc & & & & 0.60 (AUC) \\

4. Bozarth et al. \cite{bozarth-truth-2020}  & \circled{C} & Contents \textbf{(C.i)} & PolitiFact \cite{politifact}, Daily Dot, Zimdars, MBFC & LDA & \fullcirc & & & & n/a \\

5. Ciampaglia et al. \cite{ciampaglia-wiki-2015} & \circled{C} \circled{N} & Shortest path search \textbf{(C.ii)} & DBpedia & kNN, RF & \fullcirc & & & & 0.97 (AUC) \\

6. Cui et al. \cite{cui_same_2019} & \circled{C} \circled{A} \circled{U} & Content \textbf{(C.i)}; sentiment \textbf{(A.i)}; user response \textbf{(U.ii)} & PolitiFact\cite{politifact}, GossipCop\cite{gossipcop} & KNN, SVM, CSI \cite{ruchansky_csi_2017}, \underline{RMSprop} & \fullcirc & & & & 0.82 (F1) \\

7. Debnath et al. \cite{debnath_hierarchical_2020} & \circled{C} & Content \textbf{(C.i)} & LIAR \cite{wang_liar_2017} & CNN & \fullcirc & & & & 0.27 (Acc.) \\

8. Dey et al. \cite{dey_fake_2018} & \circled{C} \circled{A} & Content \textbf{(C.i)}; sentiment \textbf{(A.i)} & Twitter (API) & Clustering (kNN) & \fullcirc & & & & 0.67 (Acc.) \\

9. Galitsky et al. \cite{galitsky_detecting_2015} & \circled{C} & Content \textbf{(C.i)} & Amazon reviews & Parse thicket & \fullcirc & & & & 0.81 (Prec.)  \\

10. Glockner et al. \cite{glockner_missing_2022} & \circled{C} & Content \textbf{(C.i)} & PolitiFact\cite{politifact}, Snopes\cite{snopes}, MultiFC & CNN, DNN & \fullcirc & & & & 0.58 (Acc.) \\

11. Gordon et al. \cite{gordon_disagreement_2021} & \circled{C} \circled{A} & Content \textbf{(C.i)}; source rep \textbf{(U.i)} & Credibility-Factors2020 & SVD & \fullcirc & \fullcirc & & & 0.63 (Acc.) \\

12. Gupta et al. \cite{gupta_tweetcred_2015} & \circled{C} \circled{A} \circled{N} & Content \textbf{(C.i)}; stance \textbf{(A.iii)} & Twitter (API) & SVM & \fullcirc & \fullcirc & & & 0.60 (Agreement) \\

13. Hassan et al. \cite{hassan_data_2014} & \circled{C} & Content \textbf{(C.i)}; checkability \textbf{(C.iii)} & NBA, weather datasets & Frequency & \fullcirc & & & & n/a  \\

14. Jain et al. \cite{jain_towards_2016} & \circled{C} & Content \textbf{(C.i)}; sentiment \textbf{(A.i)} & Twitter (API) & Gensim/TextBlob & \fullcirc & & & & 0.77 (Acc.) \\

15. Jiang et al. \cite{jiang_linguistic_2018} & \circled{C}  & Content \textbf{(C.i)}; ling./syntax \textbf{(C.ii)}  & PolitiFact\cite{politifact}, Snopes\cite{snopes} & SVM & \fullcirc & & & & 0.81 (Acc.)  \\

16. Karimi et al. \cite{karimi_multi-source_2018} & \circled{C} & Content \textbf{(C.i)} & LIAR\cite{wang_liar_2017} & LSTM, CNN & \fullcirc & & & & 0.39 (Acc.)  \\

17. Kartal et al. \cite{kartal_too_2020} & \circled{C} & Content \textbf{(C.i)}; checkability \textbf{(C.iii)} & Check That! dataset & Logit, SVM, RF & \fullcirc & & & & 0.26 (MAP) \\

18. Kou et al. \cite{kou_hc-covid_2022} & \circled{C} & Content \textbf{(C.i)} & CoAID, CONSTRAINT & Knowledge graph & \fullcirc & & & & 0.90 (Acc.) \\

19. Paudel et al. \cite{paudel_lambretta_2023} & \circled{N} \circled{A} & Keyword detection \textbf{(A.ii)} & Abilov et al. dataset, Twitter (API) & AdaRank, ListNet, \underline{RF} & \fullcirc & & & & 0.79 (MAP) \\

20. Popat et al. \cite{popat_credibility_nodate} & \circled{C} & Content \textbf{(C.i)} & PolitiFact\cite{politifact}, Snopes\cite{snopes}, NewsTrust & biLSTM, CNN & \fullcirc & & & & 0.88 (AUC) \\

21. Shiralkar et al. \cite{shiralkar_finding_2017} & \circled{C} \circled{N} & KG search \textbf{(C.ii)} & DBpedia & Knowledge graph & \fullcirc & & & & 1.00 (AUC)  \\

22. Shu et al. \cite{shu_defend_2019} & \circled{C} \circled{U} & Word encoding \textbf{(C.i)}; user response \textbf{(U.ii)} & GossipCop \cite{gossipcop}, PolitiFact \cite{politifact} & \underline{RNN/RMSprop}, CSI \cite{ruchansky_csi_2017}, LSTM, CNN & \fullcirc & & \fullcirc & & 0.93 (F1) \\ 

23. Tian et al. \cite{tian_early_2020} & \circled{C} \circled{U} & Content \textbf{(C.i)}; user response \textbf{(U.ii)} & Twitter15, Twitter16 & CNN-biLSTM & \fullcirc & & \fullcirc & & 0.82 (F1) \\

24. Zhang et al. \cite{zhang_reply-aided_2019} & \circled{C} \circled{U} & Content \textbf{(C.i)}; user response \textbf{(U.ii)} & RumourEval, PHEME\cite{pheme_2018} & biLSTM, \underline{Multitask}, SVM, CNN,  & \fullcirc & \fullcirc & & & 0.89 (Acc.)  \\
\hline\hline

%%%%% end claims, begin articles %%%%%%%%%%

1. Afroz et al. \cite{afroz_detecting_2012} & \circled{A} & Content \textbf{(C.i)}; syntax \textbf{(A.i)}  & Brennan-Greenstadt & SVM, J48 Decision Trees & \fullcirc &  &  &  & 0.97 (F1) \\

2. Ahmed et al. \cite{ahmed_detecting_2018} & \circled{A}& Syntax \textbf{(A.i)} & Twitter, Kaggle, Horne and Adali \cite{horne_this_2017} & SVM & \fullcirc & & & & 0.92 (Acc.) \\

3. Bourgonje et al. \cite{bourgonje_clickbait_2017} & \circled{A} & Stance (\textbf{A.ii)} & Fake News Challenge Data & Logit & \fullcirc & & & & 0.90 (Acc.) \\

4. Brasoveanu et al. \cite{brasoveanu_semantic_2019} & \circled{A} \circled{C} & Sentiment (\textbf{A.i}); keywords \textbf{(A.ii)} & LIAR\cite{wang_liar_2017} & CNN, LSTM, \underline{CN} & \fullcirc & \fullcirc & & & 0.64 (Acc.) \\

5. Della Vedova et al. \cite{della_vedova_automatic_2018} & \circled{A} \circled{N} & Content \textbf{(C.i)}; virality \textbf{(N.iv)} &  FakeNewsNet, Buzzfeed & Logit & \fullcirc & \fullcirc & & & 0.82 (Acc.) \\

6. Horne et al. \cite{horne_this_2017} & \circled{A} & Syntax \textbf{(A.i)}, headline \textbf{(A.ii)} & Buzzfeed, Burfoot \& Baldwin & SVM & \fullcirc & & & & 0.78 (Acc.) \\

7. Jabiyev et al. \cite{jabiyev_fade_2021} & \circled{A} \circled{W} & Topic detection (\textbf{A.ii}); site cred. (\textbf{W.iv}) & Snopes\cite{snopes}, FactCheck, PolitiFact\cite{politifact}  & SVM, DT, RF & \fullcirc & & \fullcirc & & 0.87 (Acc.) \\

8. Jadhav et al. \cite{jadhav_fake_2019} & \circled{A} & Content \textbf{(C.i)}; syntax \textbf{(A.i)} & LIAR \cite{wang_liar_2017} & DSSM/RNN & \fullcirc & & & & 0.99 (Acc.)  \\

9. Jin et al. \cite{jin_detection_2017} & \circled{A} \circled{N} \circled{U}  & \textbf{(A.ii)}; \textbf{(N.i)}; suspicious accounts \textbf{(U.i)} & Tweets; articles & n/a & \fullcirc & \fullcirc & \fullcirc & & 0.87 (Prec.) \\

10. Kapusta et al. \cite{kapusta_fake_2020} & \circled{A} & Sentiment \& word freq. \textbf{(A.i)} & MBFC and custom & n/a & \fullcirc & & & & n/a  \\

11. Kumar et al. \cite{kumar_disinformation_2016} & \circled{A} \circled{W} \circled{U} & \textbf{(A.i)}; UI \textbf{(W.iii)}; Author cred. \textbf{(U.i)} & 20K Wiki Hoaxes & Random forest & \fullcirc & \fullcirc & \fullcirc & \fullcirc & 0.87 (AUC) \\

12. Magdy et al. \cite{magdy-stat-2010} & \circled{A} & Content \textbf{(C.i)} & NYT Corpus \cite{nyt_corpus}, 100 Wikis & Pattern recog. & \fullcirc & & & & 0.99 (Recall) \\

13. Monti et al. \cite{monti2019fake} & \circled{A} \circled{N} \circled{U} & \textbf{(A.ii)}; \textbf{(N.i)}; suspicious accounts \textbf{(U.i)} & Tweets; articles 
 & RNN/CNN & \fullcirc & \fullcirc & \fullcirc & & 0.927 (AUC)  \\

14. Nasir et al. \cite{nasir_fake_2021} & \circled{A} & 
Syntax \textbf{(A.i)} & ISOT \cite{isot}; FAKES \cite{fa_kes} & RNN/CNN & \fullcirc &&&& 0.99 (Acc.)  \\

15. Perez-Rosas et al. \cite{perez-rosas_automatic_2017} & \circled{A} & Syntax \textbf{(A.i)} & FakeNewsAMT; Celebrity & SVM & \fullcirc & & & & 0.74 (Acc.)\\

16. Potthast et al. \cite{potthast-etal-2018-stylometric} & \circled{A} & Syntax, sentiment, readability \textbf{(A.i)} & Buzzfeed-Webis & Bag-of-words & \fullcirc & & & & 0.46 (F1) \\

17. Reis et al. \cite{reis_explainable_2019} & \circled{A} & Syntax \textbf{(A.i)}; contents \textbf{(C.i)}; source rep. \textbf{(U.i)}; timing \textbf{(N.ii)}; URL \textbf{(W.i)} & BuzzFeed & GBM & \fullcirc & \fullcirc & \fullcirc & \fullcirc & 0.85 (AUC) \\ 

18. Rubin et al. \cite{rubin_towards_nodate} & \circled{A} & Syntax \textbf{(A.i)} & AMT & Clustering & \fullcirc & & & & 0.67 (Agreement) \\

19. Ruchansky et al. \cite{ruchansky_csi_2017} & \circled{A} \circled{U} & \textbf{(A.i)}; Responses \textbf{(A.ii)}; acct metadata \textbf{(U.i)} & Twitter/Weibo posts & RNN/LSTM & \fullcirc & & \fullcirc & & 0.95 (Acc.) \\

20. Santos et al. \cite{santos_readability_2020} & \circled{A} & Readability \textbf{(A.i)} & Fake.Br corpus & SVM & \fullcirc & & & & 0.92 (Acc.)\\ 

21. Silva et al. \cite{silva_embracing_2021} & \circled{A} \circled{N} & Topic detection (\textbf{A.ii}); propagation (\textbf{N.i}) & PolitiFact\cite{politifact}, GossipCop\cite{gossipcop}, CoAID & Clustering & \fullcirc & \fullcirc & & & 0.88 (Acc.) \\

21. Singh et al. \cite{singh_automated_nodate} & \circled{A} & Syntax \textbf{(A.i)} & Kaggle Fake News & SVM & \fullcirc & & & & 0.87 (Acc.) \\

22. Uppal et al. \cite{uppal_fake_2020} & \circled{A}  & Discourse structure \textbf{(A.i)} & Buzzfeed, PolitiFact\cite{politifact} & GRU, Dependency tree & \fullcirc & & & & 0.74 (Acc.) \\
\hline\hline

%%%%% end articles, begin users %%%%%%%%%%

1. Cao et al. \cite{cao_aiding_2012} & \circled{U} \circled{N} & Acct. cred. \textbf{(U.i)}; prop. \textbf{(N.i)} & Tuenti social network & Louvain clustering &  & \fullcirc & \fullcirc & & 0.90+ (TP) \\

2. Danezis et al. \cite{danezis2009sybilinfer} & \circled{U} \circled{N} & Acct. cred. \textbf{(U.i)}; prop. \textbf{(N.i)} & LiveJournal data & Bayesian inf. & & \fullcirc & \fullcirc & & n/a*  \\

3. Ezzeddine et al. \cite{ezzeddine2023exposing} & \circled{U} & Acct. behaviors \textbf{(U.ii)} & DATA & LSTM & & \fullcirc & \fullcirc & & 0.91 (AUC) \\

4. Hamdi et al. \cite{hamdi_hybrid_2020} & \circled{U} \circled{N} & Account metadata \textbf{(U.i)}; prop. \textbf{(N.i)} & CREDBANK & LDA, Bayes, Logit, SVM & & \fullcirc & \fullcirc & & 0.99 (AUC) \\

5. Helmstetter et al. \cite{helmstetter-supervised-2018} & \circled{U} \circled{N} \circled{A} & Acct metadata \textbf{(U.i)}; post sharing data \textbf{(U.ii)} & Public site cred. lists & SVM, NB, DT, RF & \fullcirc & \fullcirc & \fullcirc & & 0.936 (F1) \\

6. Jain et al. \cite{jain_towards_2016} & \circled{U} \circled{A} & Acct metadata \textbf{(U.i)}; topic det. \textbf{(A.ii)} & Twitter (API) & Gensim, TextBlob & \fullcirc & & \fullcirc & & 0.77 (Acc.) \\

% 4. Helmstetter et al. \cite{helmstetter-supervised-2018} & \circled{U} & 0.982 \\

7. Leonardi et al. \cite{leonardi_users_2021} & \circled{U} \circled{N} \circled{A} & Acct metadata \textbf{(U.i)}; prop. \textbf{(N.i)} & CoAID & RF & \fullcirc & \fullcirc & \fullcirc & & 0.81 (F1) \\

8. Saeed et al. \cite{saeed2022trollmagnifier} & \circled{U} &  User behavior \textbf{(U.ii)} & Reddit Pushshift; Reddit IRA trolls list & RF & & & \fullcirc & & 0.98 (Acc.) \\

9. Sansonetti et al. \cite{sansonetti_unreliable_2020} & \circled{U} \circled{C} & Acct metadata \textbf{(U.i)}; acct activity \textbf{(U.ii)}  & PolitiFact \cite{politifact}, Twitter (API) & \underline{LSTM-CNN}, SVM, KNN & \fullcirc & & \fullcirc & & 0.92 (Acc.) \\

10. Santia et al. \cite{santia_bots_2019} & \circled{U} \circled{A} & Source rep. \textbf{(U.i)}; user response \textbf{(U.ii)}; syntax \textbf{(A.i)} & BuzzFeed & SVM, \underline{RF}, DT, NB & \fullcirc & \fullcirc & \fullcirc & & 0.77 (Prec.) \\

11. Shu et al. \cite{shu_exploiting_2017} & \circled{U} \circled{N} \circled{A} & User behavior \textbf{(U.ii)}; prop. \textbf{(N.i)} syntax \textbf{(A.i)} & Buzzfeed, PolitiFact & Gibbs sampling & \fullcirc & \fullcirc & \fullcirc & & 0.85+ (Acc.)\\

12. Vargas et al. \cite{vargas2020detection} & \circled{N} \circled{A} & Prop. \textbf{(N.i)}; topic det. \textbf{(A.ii)} & Twitter (API) & RF & \fullcirc & \fullcirc & & & 0.98 (F1) \\

13. Wang et al. \cite{wang_sybil_2013} & \circled{U} \circled{N} & User behavior \textbf{(U.ii)} & Renren data & SVM & & \fullcirc & \fullcirc & & 0.99 (Acc.) \\

14. Yu et al. \cite{yu_sybillimit_2008} & \circled{U} \circled{N} & User behavior \textbf{(U.i)}; prop. \textbf{(N.i)} & LiveJournal, Friendster, DBLP accounts & Random route & & \fullcirc & \fullcirc & &  n/a* \\

15. Yuan et al. \cite{yuan2019detecting} & \circled{U} \circled{N} & Acct metadata \textbf{(U.i)}; timing \textbf{(N.ii)} & WeChat data & Clustering & & \fullcirc & \fullcirc & & 0.90+ (Prec.) \\

16. Zhang et al. \cite{zhang_misinformation_2016} & \circled{U} \circled{N} & User behavior \textbf{(U.ii)}; prop. \textbf{(N.i)} & Twitter, Slashdot, Epinion & Graph cut & & \fullcirc & \fullcirc & & n/a \\

17. Zhou et al. \cite{zhou_network-based_2019} & \circled{U} \circled{N} & User suscept. \textbf{(U.i)}; prop \textbf{(N.i)} & PolitiFact \cite{politifact}, BuzzFeed & SVM, KNN, NB, DT, \underline{RF} & & \fullcirc & \fullcirc & & 0.93 (Acc.) \\
\hline\hline

%%%%% end users, begin networks %%%%%%%%%%

1. Alizadeh et al. \cite{alizadeh2020content} & \circled{N} \circled{A} \circled{U} & Propagation \textbf{(N.i)}; syntax \textbf{(A.i)}; acct metadata \textbf{(U.i)} & Twitter (API), Reddit IRA troll list & RF & \fullcirc & \fullcirc & \fullcirc & & 0.70+ (F1) \\

2. Antoniadis et al. \cite{antoniadis_model_2015} & \circled{U} \circled{A}  & Acct metadata \textbf{(U.i)}; syntax \textbf{(A.i)} & Hurricane Sandy tweet dataset & J48, \underline{RF}, KNN, Bayes & \fullcirc & & \fullcirc & & 0.79 (Avg. Prec.) \\

3. Assenmacher et al. \cite{assenmacher2020two} & \circled{N} \circled{A} & Propagation \textbf{(N.i)}; topic det. \textbf{(A.ii)} & Twitter (API) & Clustering & \fullcirc & \fullcirc & & & not reported \\

4. Buntain et al. \cite{buntain_automatically_2017}  & \circled{N} \circled{U} \circled{A} & Time \textbf{(N.ii)}; acct metadata \textbf{(U.i)}; sentiment \textbf{(A.i)} & CREDBANK, Buzzfeed & RF & \fullcirc & \fullcirc & \fullcirc & & 0.65 (Acc.) \\

% 5. Cao et al. \cite{cao_aiding_2012} & \circled{N} & -- \\

5. Castillo et al. \cite{castillo_information_2011} & \circled{U} \circled{A} & Syntax \textbf{(A.i)}; user behavior \textbf{(U.ii)} & Twitter Monitor events & SVM, DT & \fullcirc & \fullcirc & \fullcirc & & 0.874 (P) \\

6. Chen et al. \cite{chen_rumor_2018} & \circled{N} \circled{U} \circled{A} & Social graph \textbf{(N.iii)}; syntax \textbf{(A.i)}; user behavior \textbf{(U.ii)} & Weibo & RNN & \fullcirc & \fullcirc & \fullcirc & & 0.92 (Acc.) \\

7. Guo et al. \cite{guo_rumor_2018} & \circled{N} \circled{C} & Prop. \textbf{(N.i)}; semantics \textbf{(C.i)} & Twitter, Weibo & LSTM & \fullcirc & \fullcirc & & & 0.9 (Acc.) \\

8. Jin et al. \cite{jin_news_2016} & \circled{N} \circled{A} & Propagation \textbf{(N.i)}; stance \textbf{(A.iii)} & Sina Weibo posts & Clustering & \fullcirc & \fullcirc & & & 0.84 (Acc.) \\

9. Liu et al. \cite{liu_early_nodate} & \circled{N} & Propagation \textbf{(N.i)} & Weibo, Twitter15, Twitter16 & RNN, CNN & & \fullcirc & & & 0.897 (Acc.) \\

10. Ma et al. \cite{ma_detect_2017} & \circled{N} & Propagation \textbf{(N.i)} & Kochina, Ma, Shu Twitter datasets & RNN/BiLSTM & & \fullcirc & & & 0.75 (Acc.) \\

11. Magelinski et al. \cite{magelinski_synch_2022} & \circled{N} \circled{U} & Prop. \textbf{(N.i)}; timing \textbf{(N.ii)}; user behavior \textbf{(U.ii)} & Twitter (API) & -- & & \fullcirc & \fullcirc & & -- n/a \\

12. Nguyen et al. \cite{nguyen_fake_2019} & \circled{N} \circled{A} & Prop. \textbf{(N.i)}; semantics \textbf{(A.i)} & Twitter, Weibo, PHEME \cite{pheme_2018} & SVM, RNN, DT & \fullcirc & \fullcirc &  &  & 0.970 (Acc.) \\

13. Pacheco et al. \cite{pacheco_uncovering_2020} & \circled{U} \circled{N} & Account metadata \textbf{(U.i)}; timing \textbf{(N.i)} & Twitter (API) & Clustering &  & \fullcirc & \fullcirc & & 0.8+ (Prec.) \\

14. Ratkiewicz et al. \cite{ratkiewicz2011detecting} & \circled{N} \circled{A} \circled{U} & Prop. \textbf{(N.i)}; keywords \textbf{(A.ii)}; user behavior \textbf{(U.ii)} & Twitter (API) & AdaBoost, \underline{SVM} & \fullcirc & \fullcirc & \fullcirc & & 0.96 (Acc.) \\

15. Sharma et al. \cite{sharma_identifying_2021} & \circled{N} \circled{A} \circled{U} & Prop. \textbf{(N.i)}; keywords \textbf{(A.ii)}; user behavior \textbf{(U.ii)} & Twitter (API); Twitter IRA trolls list & GMM, \underline{Kmeans}, NN &  \fullcirc & \fullcirc & \fullcirc & & 0.94 (AUC) \\

16. Tschiatschek et al. \cite{tschiatschek_fake_2018} & \circled{N} \circled{U} & Prop. \textbf{(N.i)}; user rep. \textbf{(U.i)} & Facebook dataset & Bayes inf. & & \fullcirc & \fullcirc & & n/a \\

17. Zeng et al. \cite{zeng_unconfirmed_2016} & \circled{N} \circled{A} & Prop. \textbf{(N.i)}; stance \textbf{(A.iii)} & 4.3K tweets (from API) & Logit, NB, RF & \fullcirc & \fullcirc & & & 0.88 (Acc.) \\

\hline\hline
%%%%% end networks, begin websites %%%%%%%%%%
1. Asr et al. \cite{torabi_asr_data_2018} & \circled{W} \circled{A} & Source rep. \textbf{(W.i)}; Syntax \textbf{(A.i)} & BuzzfeedUSE, Snopes, Rashkin, Rubin & CNN, \underline{SVM}, NB & \fullcirc & \fullcirc &  & \fullcirc & not reported \\
2. Baly et al. \cite{baly_predicting_nodate} & \circled{W} \circled{A} \circled{N} & Source rep. \textbf{(W.i)}; Site infra. \textbf{(W.ii)}; \textbf{(A.i)} & MediaBiasFactCheck\cite{mediabiasfactheck} &  SVM & \fullcirc & & & \fullcirc & 0.66 (Acc.) \\ 
3. Baly et al. \cite{baly-etal-2020-written} & \circled{W} \circled{A} \circled{N} & Source rep. \textbf{(W.i)}; Site infra. \textbf{(W.ii)}; \textbf{(A.i)} & MediaBiasFactCheck\cite{mediabiasfactheck} & SVM & \fullcirc & & & \fullcirc & 0.7152 (Acc.) \\ 
4. Castelo et al. \cite{castelo_agnostic_2019} & \circled{W} \circled{A} & Site infra. \textbf{(W.ii)}; syntax \textbf{(A.i)} & Celebrity, US-Election2016 & \underline{SVM}, kNN, RF & \fullcirc & & & \fullcirc & 0.86 (Acc.) \\
5. Chen et al. \cite{chen_proactive_2020} & \circled{W} \circled{A} & Hosting infra. (URL) \textbf{(W.ii)}; syntax \textbf{(A.i)} & PoliticalFakeNews & Clustering & \fullcirc & & & \fullcirc & 0.97 (AUC) \\
6. Hounsel et al. \cite{hounsel-infra-2020} & \circled{W} & Site infra. \textbf{(W.i)} & FactCheck, Snopes, PolitiFact, Buzzfeed & RF & & & & \fullcirc & 0.98 (AUC) \\
7. Ribeiro et al. \cite{ribeiro_media_2018} & \circled{W} \circled{U} & Site infra. \textbf{(W.i)}; User demog. \textbf{(U.i)} &  Facebook API & Graph search & & \fullcirc & \fullcirc & \fullcirc & 0.97 (PCC) \\
    \end{tabular}
}
\caption{\textbf{Focus corpus by scope and target.} Coding of a focus set of 87 papers, sorted by information scope. Values in parentheses in ``Target'' field correspond to highlighted subcategories presented in~\Cref{sec:systematization_of_literature} (e.g., ``C.i'' denotes target (i) in ``Claims,''~\Cref{sec:claims}). If authors present evaluation results for multiple models, we \underline{underline} the most performant model and record its corresponding performance score.}
\label{app:full_coding}
\end{table*}

\begin{table*}
\footnotesize
    \def\arraystretch{1.4}
    \centering
        \begin{tabular}{lc|cc|cc|cc|cc|cc} 
        \toprule 
%     \hline
Dataset (size) & All features & \multicolumn{2}{p{2cm}}{\centering \texttt{articles} \\ $-$ \hspace{0.75cm} $+$} & \multicolumn{2}{p{2cm}}{\centering \texttt{traffic} \\ $-$\hspace{0.75cm} $+$} &  \multicolumn{2}{p{2cm}}{\centering \texttt{twitter} \\ $-$\hspace{0.75cm} $+$} & \multicolumn{2}{p{2cm}}{\centering \texttt{wikipedia} \\ $-$\hspace{0.75cm} $+$} & \multicolumn{2}{p{2cm}}{\centering \texttt{url} \\ $-$\hspace{0.75cm} $+$} \\
        \midrule
            Full corpus (1066)  & 0.654 & \cellcolor{red!25}0.631 & \cellcolor{teal!70}0.644 & 0.654 & 0.508 & 0.648 & 0.550 & \cellcolor{red!70}0.627 & \cellcolor{teal!20}0.606 & 0.638 & 0.533\\ 
            Med. corpus (400)   & 0.623 & \cellcolor{red!25}0.608 & \cellcolor{teal!70}0.630 & 0.620 & 0.488 & 0.635 & 0.500 & \cellcolor{red!70}0.590 & \cellcolor{teal!20}0.588 & 0.623 & 0.495\\
            Small corpus (250)  & 0.636 & 0.632 & \cellcolor{teal!70}0.596 & 0.632 & 0.524 & \cellcolor{red!25}0.608 & 0.512 & \cellcolor{red!70}0.588 & \cellcolor{teal!20}0.536 & 0.624 & 0.516 \\
            Left bias (398)     & 0.691 & \cellcolor{red!25}0.683 & \cellcolor{teal!20}0.671 & 0.688 & 0.668 & 0.686 & 0.628 & \cellcolor{red!70}0.678 & \cellcolor{teal!70}0.683 & \cellcolor{red!70}0.678 & 0.636\\
            Center (263)   & 0.913 & \cellcolor{red!70}0.810 & \cellcolor{teal!70}0.890 & 0.913 & 0.700 & 0.924 & 0.741 & 0.920 & \cellcolor{teal!20}0.776 & \cellcolor{red!35}0.890 & 0.635\\
            Right bias (405)    & 0.279 & \cellcolor{red!70}0.267 & \cellcolor{teal!70}0.252 & 0.279 & 0.173 & 0.286 & \cellcolor{teal!20}0.230 & 0.274 & 0.205 & \cellcolor{red!35}0.272 & 0.121\\
            \midrule
            Full corpus (1066)  & 0.569 & \cellcolor{red!70}0.523 & \cellcolor{teal!70}0.595 & 0.569 & 0.399 & 0.580 & 0.440 & \cellcolor{red!35}0.552 & 0.538 & 0.577 & 0.373\\ 
            Med. corpus (400)   & 0.563 & \cellcolor{red!70}0.517 & \cellcolor{teal!70}0.580 & \cellcolor{red!35}0.560 & 0.420 & 0.578 & 0.478 & 0.585 & 0.545 & 0.568 & 0.360\\
            Small corpus (250)  & 0.456 & \cellcolor{red!35}0.424 & \cellcolor{teal!70}0.560 & 0.452 & 0.364 & 0.500 & 0.408 & \cellcolor{red!70}0.400 & 0.496 & 0.444 & 0.436 \\
            Low cred. (256)     & 0.590 & \cellcolor{red!70}0.516 & \cellcolor{teal!20}0.633 & \cellcolor{red!35}0.590 & \cellcolor{teal!70}0.641 & 0.629 & 0.445 & 0.609 & 0.633 & \cellcolor{red!35}0.590 & 0.473\\
            Mixed cred. (268)   & 0.407 & \cellcolor{red!70}0.340 & \cellcolor{teal!70}0.474 & 0.407 & 0.0522 & 0.414 & 0.258 & \cellcolor{red!35}0.362 & \cellcolor{teal!20}0.276 & 0.414 & 0.198\\
            High cred. (542)    & 0.349 & \cellcolor{red!70}0.336 & \cellcolor{teal!70}0.408 & 0.349 & 0.255 & 0.369 & 0.271 & \cellcolor{red!35}0.341 & \cellcolor{teal!20}0.316 & 0.351 & 0.218\\
        \bottomrule
        \end{tabular}
    \caption{Replication analysis of Baly et al.: Dropout(--) and feature importance(+) analyses of subsets of Baly et al.'s EMNLP18 dataset, stratified by political leaning and credibility. Most (secondmost) performant feature, as determined by its contribution to overall classifier accuracy on the full feature set, is highlighted in darker (lighter) hues. Fact and bias classification task performances are reported in the top and bottom halves of the table, respectively. }
    \label{tab:baly_ablation}
\end{table*}

%%%%%%%%%%%%%%%%%%%%%%%%%%%%%%%%%%%%%%%%%%%%%%%%%%%%%%%%%%%%%%%%%%%%%%%%%%%%%%%%
\end{document}